\documentclass[letterpaper, 10 pt, conference]{ieeeconf} 
\IEEEoverridecommandlockouts
\usepackage{cite}
\usepackage{amsmath,amssymb,amsfonts}
\usepackage{algorithmic}
\usepackage{graphicx}
\usepackage{textcomp}
\usepackage{xcolor}

\usepackage{hyperref}       
\usepackage{url}            
\usepackage{booktabs}       
\usepackage{amsbsy}
\usepackage{nicefrac}       
\usepackage{subcaption}
\usepackage{datetime}
\usepackage{float}
\usepackage{algorithm}
\usepackage{graphics}
\usepackage{multirow}
\usepackage{xcolor}
\usepackage{mathtools}

\usepackage{xspace}
\usepackage[subtle,tracking=normal]{savetrees}

\begin{document}

\title{Representation Matters: Improving Perception \\and Exploration for Robotics
}

\author{
Markus Wulfmeier,
Arunkumar Byravan,
Tim Hertweck,
Irina Higgins,
Ankush Gupta,
Tejas Kulkarni,\\
Malcolm Reynolds,
Denis Teplyashin,
Roland Hafner,
Thomas Lampe,
Martin Riedmiller
\thanks{DeepMind, London, United Kingdom}
\thanks{Correspondence author:
        {\texttt{mwulfmeier@google.com}.}}
}

\maketitle

\begin{abstract}
Projecting high-dimensional environment observations into lower-dimensional structured representations can considerably improve data-efficiency for reinforcement learning in domains with limited data such as robotics.
Can a single generally useful representation be found? In order to answer this question, it is important to understand how the representation will be used by the agent and what properties such a ‘good’ representation should have. 
In this paper we systematically evaluate a number of common learnt and hand-engineered representations in the context of three robotics tasks: lifting, stacking and pushing of 3D blocks. The representations are evaluated in two use-cases: as input to the agent, or as a source of auxiliary tasks. Furthermore, the value of each representation is evaluated in terms of three properties: dimensionality, observability and disentanglement.
We can significantly improve performance in both use-cases and demonstrate that some representations can perform commensurate to simulator states as agent inputs.
Finally, our results challenge common intuitions by demonstrating that: 
1) dimensionality strongly matters for task generation, but is negligible for inputs, 2)  observability of task-relevant aspects mostly affects the input representation use-case, and 3) disentanglement leads to better auxiliary tasks, but has only limited benefits for input representations.
This work serves as a step towards a more systematic understanding of what makes a ‘good’ representation for control in robotics, enabling practitioners to make more informed choices for developing new learned or hand-engineered representations.
\end{abstract}

\section{Introduction}\label{sec:intro}

Many reinforcement learning (RL) domains, including the real world, are naturally accessed via high-dimensional, unstructured observations, such as images. The underlying world state, however, can be represented using more compressed structured representations according to the widely held manifold hypothesis. Such structured representations can potentially simplify the learning of downstream tasks. However, a good mapping from observations to representations is in general not available and needs to be learnt. Many representation learning approaches have been proposed in the past to address this challenge, using reconstruction-based \cite{hinton2006reducing, kingma2013auto, higgins2017beta, gregor2016towards, burgess2019monet, kulkarni2019unsupervised} or more recently, contrastive objectives \cite{oord2018representation,chen2020simple, srinivas2020curl, grill2020bootstrap} in the context of deep learning. Different approaches result in representations with different properties, but systematic investigations on how these properties affect their utility for improving performance on downstream RL tasks have been limited. 
The majority of past work has concentrated on one particular application of representations in the RL context---as a transformation of high dimensional observations into low-dimensional state inputs to the agent policy or value function \cite{higgins2017darla,tassa2018deepmind,kaiser2019model}. 
However, representations may also be used in other ways, for example as a source of intrinsically generated tasks to guide exploration. 
Well-structured manually-designed auxiliary tasks have been shown to demonstrate considerable success to accelerate learning in multitask and transfer learning in the robotics environment \cite{riedmiller2018learning,wulfmeier2019regularized,teh2017distral}, a domain where data-efficiency is of utmost importance. If representation learning can help automate the process of auxiliary task design, it can be a significant step change.
Past work has touched on this, for individual hand-crafted \cite{sharma2019dynamics,hertweck2020simple} or learnt representations \cite{grimm2019disentangled,relevantfreatures20,laversannefinot2018curiosity,nair2018visual}. However, no systematic investigation has been performed to compare different representations as sources of auxiliary tasks and to analyze the relation between use-cases and different properties.

\begin{figure}[t]
	\centering
	\begin{tabular}{c}
	\hspace{-3mm}
        \includegraphics[width = 0.48\textwidth]{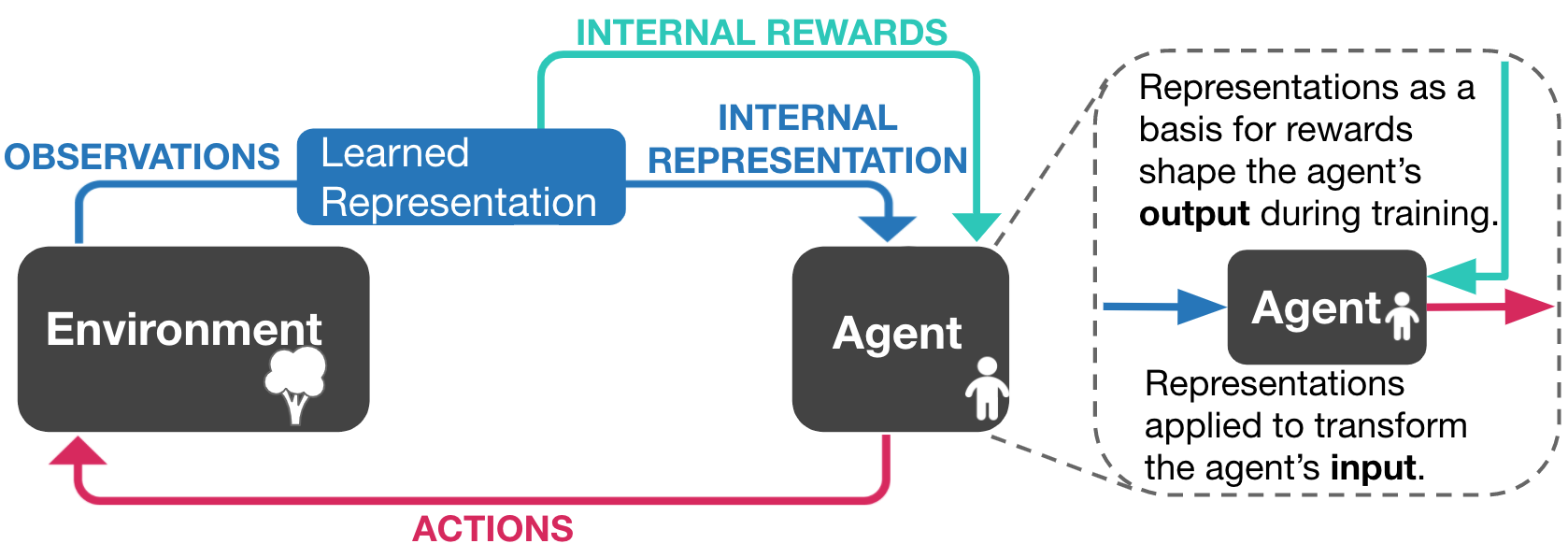}
	\end{tabular}
    \caption{Principal use-cases for representation learning in agents investigated in this work. External reward omitted for clarity.}
	\label{fig:main}
	\vspace{-3mm}
\end{figure}

In this paper, we evaluate the role of representation learning for improving the efficiency of learning in the robotics domain. 
We obtain the representations using a number of promising modern deep representation learning approaches: MONET \cite{burgess2019monet}, Transporter \cite{kulkarni2019unsupervised}, and $\beta-$VAE \cite{higgins2017beta}. These representations vary in dimensionality, level of disentanglement (whether the representation can be decomposed into a product of semantically interpretable independent sub-spaces \cite{higgins2018towards}), as well as observability. We compare these approaches to a number of baselines, such as random projection and hand-engineered feature extraction \cite{hertweck2020simple}, on three complex, sparse reward tasks---lifting, stacking and pushing objects to match the target position, in a simulated 3D robotics manipulation environment. We demonstrate that well chosen representations can significantly accelerate learning in both use-cases: as input representations and as sources of auxiliary tasks. In addition, well chosen representations, such as MONET \cite{burgess2019monet} and Transporter \cite{kulkarni2019unsupervised}, can perform on par with ground-truth simulator states - essentially bridging the gap between state and pixel-based reinforcement learning.

Our main contributions include:
\begin{itemize}
    \item Considerably improved agent performance in three simulated robotics tasks via two different representation use-cases. The application of MONET and Transporter to transform input images leads to performance on par with ground-truth simulator states.
    \item Detailed investigation of common learnt representations with respect to three general properties and resulting insights for using and developing representation models.
    \item Improved task scheduling for addressing challenges 
    from representation dimensionality and larger task sets.
\end{itemize}

Our results extend common intuitions by demonstrating that:
\emph{Dimensionality} strongly matters in the auxiliary task use-case, with higher-dimensional representations producing more auxiliary tasks and putting extra burden on the multi-task learning system, but has negligible effect when used as inputs. 
\emph{Observability} of task-relevant aspects affects the agent in the representation-as-input use-case. The loss of important information in the input to the agent results in partial observability of the environment and renders certain tasks impossible to solve. 
Higher \emph{disentanglement} leads to better auxiliary tasks, since the representation can be naturally decomposed into independently controllable features \cite{grimm2019disentangled,bengio2017independently}. On the other hand, it has limited impact when used as inputs with the exception of benefits of interpretability of learnt representations which enable to manually extract lower-dimensional representations that contain all the information necessary to solve the task.
\section{Methods}\label{sec:methods}
We introduce all necessary reinforcement and representation learning formalism in this section and conclude with a description of the two use-cases.

\subsection{Multi-Task Reinforcement Learning}\label{sec:multi}
The problem of multi-task reinforcement learning (MTRL) can be represented as a set of Markov Decision Processes (MDPs), that share all aspects with the exception of task specific information.
The MDP formalism describes a tuple $(S, A, p, r, \gamma)$ with the sets of states and actions, respectively $S$ and $A$, the transition probability $p(s^\prime|s,a)$, the reward $r(s,a)$ and the discount factor $\gamma$. Now, in MTRL the MDPs share all properties except that each has its own reward function $r_i$ with $i \in I$ indexing into the set of tasks. 
In the general case, we are unable to access the underlying environment state and instead perceive the observation $x$ modelled by the set of observations $O$ and the conditional distribution $x \sim o(\cdot|s)$.
The agent for each task $i$ acts according to its policy $\pi_i(a|x)$, which we optimize off-policy \cite{abdolmaleki2018relative} to maximize the agent's objective $J(\pi) =
\mathbb{E}_{i \sim I}\Big[
\mathbb{E}_{\pi_i, p\left(s_0\right)}\Big[\sum_{t=0}^{\infty} \gamma^{t} r_i\left(s_{t}, a_{t}\right) | s_{t+1} \sim p(\cdot|s_t,a_t) \Big]\Big].$

\begin{figure}[t]
\hspace{-4mm}
\begin{minipage}[b]{.33\linewidth}
\centering

\begin{tabular}{c} 
    \includegraphics[trim=0cm 6.8cm 4.5cm 0,clip,width = .98\textwidth]{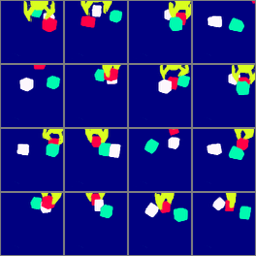}
\end{tabular}
\end{minipage}
\begin{minipage}[b]{.33\linewidth}
\hspace{-0mm}
\begin{tabular}{cc}
    \includegraphics[width = 0.45\textwidth,height = 0.49\textwidth]{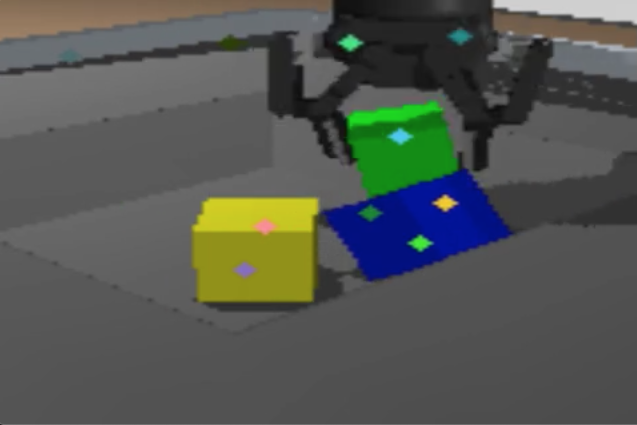}\hspace{-1mm}
    \includegraphics[width = 0.45\textwidth,height = 0.49\textwidth]{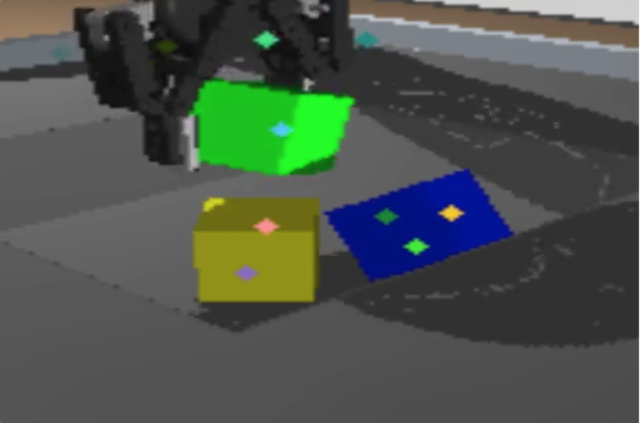} 
\end{tabular}
\end{minipage}
\begin{minipage}[b]{.33\linewidth}
\hspace{-4.6mm}
\begin{tabular}{cc}
    \includegraphics[width = 0.49\textwidth]{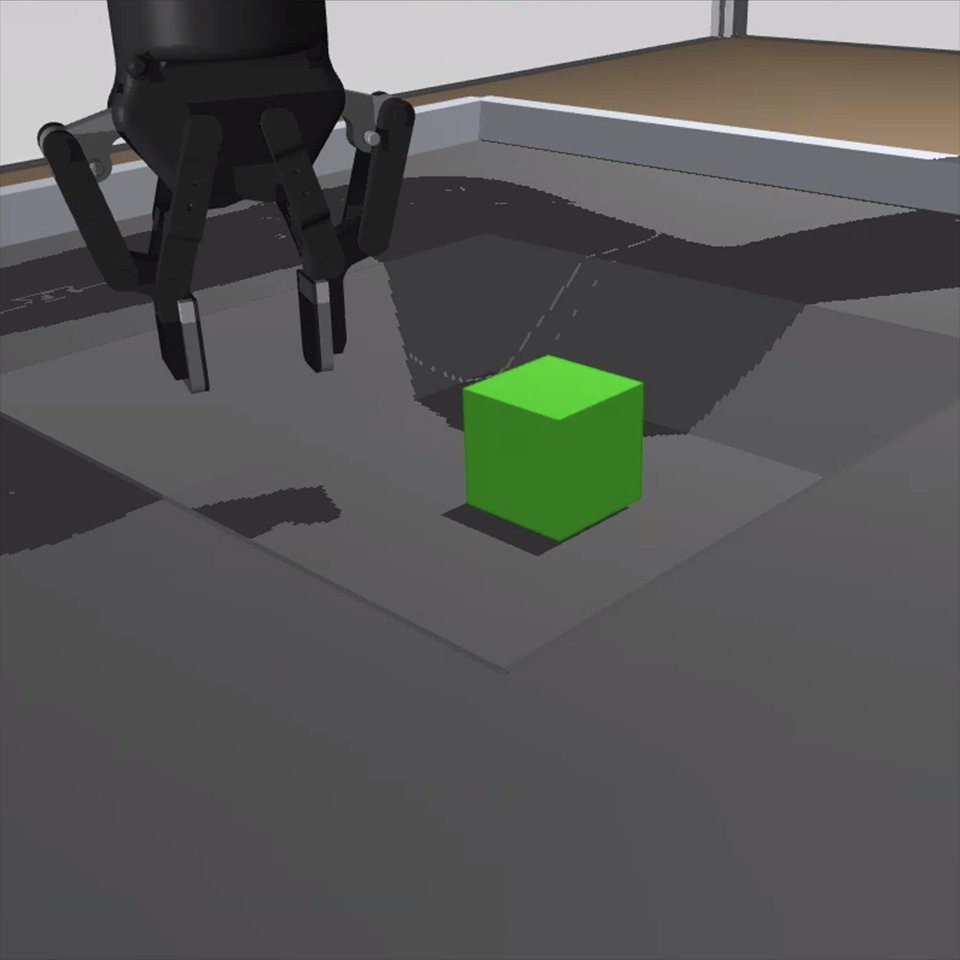}\hspace{-1mm}
    \includegraphics[width = 0.49\textwidth]{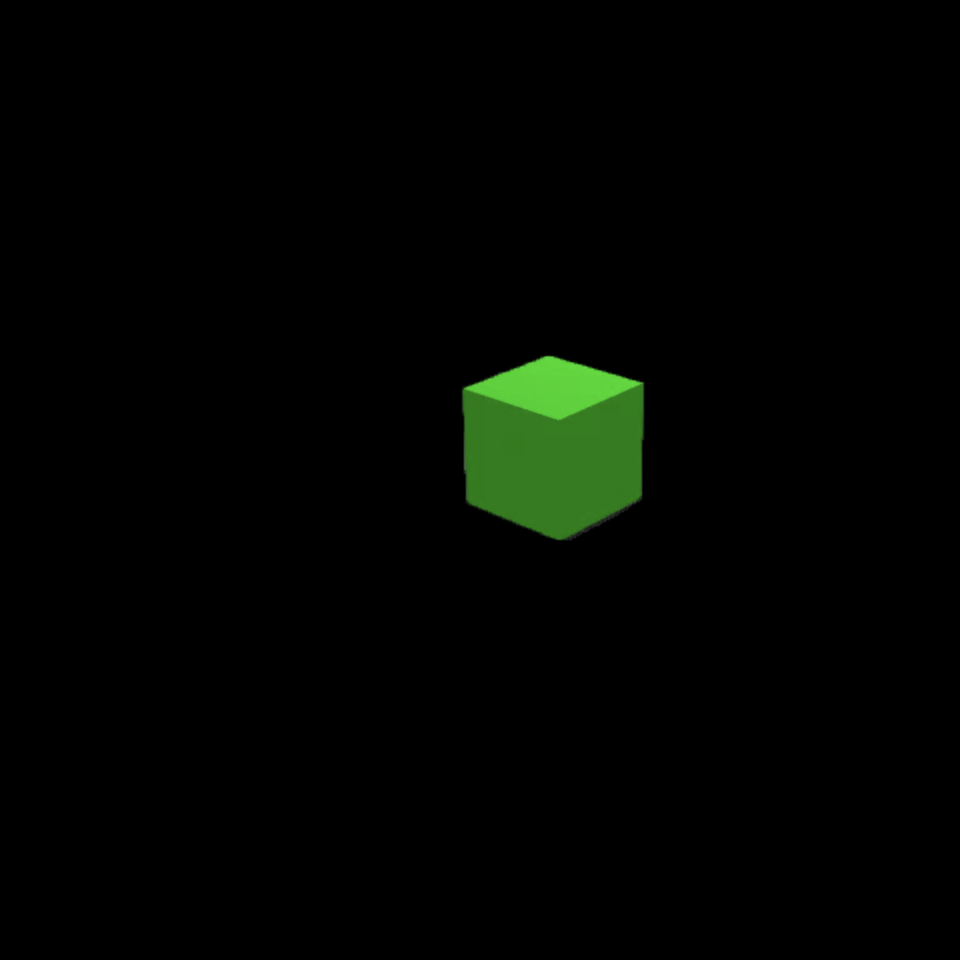}
\end{tabular}
\end{minipage}
\caption{Visualizations of learned unsupervised representations. Left: Object segmentations learned from MONet, Middle: Keypoints from Transporter projected onto the image, Right: Original observation and hand defined color segmentation.}
\label{fig:representations}
\vspace{-3mm}
\end{figure}

\subsection{Representation Learning}\label{sec:rep}
High-dimensional observations, such as images, can increase the difficulty of RL and reduce sample efficiency \cite{tassa2018deepmind, kaiser2019model}. In order to render learning more tractable, we often aim to obtain intermediate lower-dimensional state representations. In this paper, we explore the use of several recent representation learning methods that map a high-dimensional image observation $I$ to an embedding $z$; this embedding can be used as an input to the RL agent instead of the pixel observations (Sec.~\ref{sec:input}), or to specify downstream tasks/rewards (Sec.~\ref{sec:tasks}). 
Fig.~\ref{fig:representations} visualizes some of the learned representations in our simulated environment \footnote{We provide videos and further method and experiment details including network architectures and hyper-parameters in the appendix under \url{https://sites.google.com/view/representation-matters}}.  

\paragraph{Transporter}\label{sec:transporter}
Transporter~\cite{kulkarni2019unsupervised} is an unsupervised method for discovering \emph{keypoints} of independently moving entities in the environment. Temporally consistent keypoint detectors are learned by sampling pairs of frames $I, I^\prime$, where the target frame $I^\prime$ is reconstructed by transporting features at the learned keypoint locations from the source frame $I$. The embedding $z\in\mathbb{R}^{K{\times}2}$ corresponds to these learned object coordinates, where each of the $K$ objects (this number is pre-specified) has 2 co-ordinates in the image frame.

\paragraph{($\beta$-) Variational Autoencoder}\label{sec:vae}
A Variational Autoencoder (VAE) \cite{rezende2014stochastic,kingma2013auto} is a generative model that aims to learn a joint distribution $p(I, Z)$ of pixel images $I$ and a set of latent variables $Z$ by learning to maximize the evidence lower bound on the data distribution. VAEs use an encoder that takes in images $I$ and parameterizes the posterior distribution $q(Z|I)$, and a decoder that reconstructs the image $I$ from sampled latents $\hat{z}$. They are trained by minimizing the sum of the reconstruction loss and a compression term that encourages the posterior $q(Z|I)$ to be close to a prior term. A $\beta$-VAE \cite{higgins2017beta} is a variation of the VAE model, where the compression term is scaled by a $\beta$ hyperparameter; this controls the degree of disentangling achieved by the model. In this work, we train a model with a low $\beta$ when we aim for an entangled representation and a high $\beta$ when we aim for a disentangled representation. We use the mean of the learned posterior as the embedding $z$.

\paragraph{MONet}\label{sec:monet}
MONet \cite{burgess2019monet} is an unsupervised method for discovering object-based representations of visual scenes. It augments a VAE (or a $\beta$-VAE) with a recurrent attention network, which partitions the image into $N$ attention masks, each of which corresponds to a meaningful entity in the visual scene (e.g. an object, or the background). These attention attenuated images are then fed into a shared VAE applied sequentially to them, and the weighted sum of the respective reconstructions is trained to correspond to the original input image. The embedding $z$ is the concatenation of the $N$ embeddings obtained from the sequential application of the VAE to the $N$ attention-attenuated input images.

\paragraph{Color Segmentation}\label{sec:color}

In \cite{hertweck2020simple} the authors provide a method for deriving a simple, high-level representation by aggregating statistics of an image’s spatial color distribution. Partitioning the hue range into multiple brackets and thresholding returns a color-based segmentation. We compute the mean and variance (in pixel space) along the image axes of these segmentation masks and concatenate these across chosen colors.

\paragraph{Random Projection}\label{sec:random}
As a data-independent dimensionality reduction method, we also investigate a simple random projection approach \cite{bingham2001random}. We randomly initialize a matrix $M \in \mathbb{R}^{(h*w*3,d)}$ where $h$ and $w$ denote the image height and width respectively and $d$ is the embedding size. The embedding $z$ can be computed as $z = \text{tanh} (M \cdot \text{flatten}(I)) $, where we apply a saturating non-linearity (tanh) for bounding the output. 

\subsection{Representations for Input Transformation}\label{sec:input}
When using the representations to transform agent inputs, we use the models as transformations such that both policy and Q-function are conditioned on the embedding instead of the full state, respectively as $\pi(a|z(x))$ and $Q(z(x),a)$. 

\subsection{Representations for Auxiliary Tasks}\label{sec:tasks}
There are many possible ways to define tasks based on a structured representation of the environment. We choose a simple approach where for each dimension $i$ of the embedding $z(x)$, we create two tasks, denoted $min$ and $max$, respectively for minimizing and maximizing its value. 
These rewards can be assigned automatically by running inference on the representation learning models with trajectories generated from any task and we can randomly sample tasks for every update. Further details on auxiliary task generation are in the appendix. 

\paragraph{Off-Policy Learning and Hindsight Reward Assignment}\label{sec:hindsight}
In on-policy learning, each policy can only be updated from data that it has just created. This means we cannot share data even when training for multiple tasks for the same agent embodiment and domain. However, this problem is circumvented in off-policy learning, where we can e.g. re-weight data based on importance sampling accounting for the difference between the data-generating and data-fitting policies \cite{sutton1998introduction}. 
In this way, off-policy learning enables us to share data across tasks given the ability to generate rewards for all tasks in hindsight \cite{andrychowicz2017hindsight, riedmiller2018learning, nachum2018data}; we take advantage of this in our work. 

\paragraph{Task Scheduling}\label{sec:scheduling}
In multi-task learning the most common approach is to randomly sample a new task for each episode with the corresponding agent policy and reward function. 
When providing large sets of tasks to our agent, we build on two mechanisms for further improvement of robustness and efficiency: 1. learning a model for task choice; 2. choosing multiple tasks to run in sequence during a single episode. The former enables us to focus on features relevant to a final task, and the latter enables to train robust policies and help exploration by having each task affect other tasks' initial state distributions \cite{riedmiller2018learning}.
We build upon the Q-scheduler \cite{riedmiller2018learning} which has two independent phases for scheduling. In the first phase, when no reward for the main task has been observed, we use a purely random schedule. After we begin to observe rewards for the main task, scheduling changes to increase the preference of tasks proportional to the main task rewards their trajectories generate. 
We improve the initial phase by replacing random scheduling with a preference for tasks which currently generate rewards but are not fully solved yet; this provides a strong learning signal throughout training.

\section{Experiments}\label{sec:experiments}

\begin{figure}[t]
\begin{minipage}[b]{\linewidth}
    \centering
    \begin{tabular}{cccc}
    \includegraphics[height=0.28\textwidth]{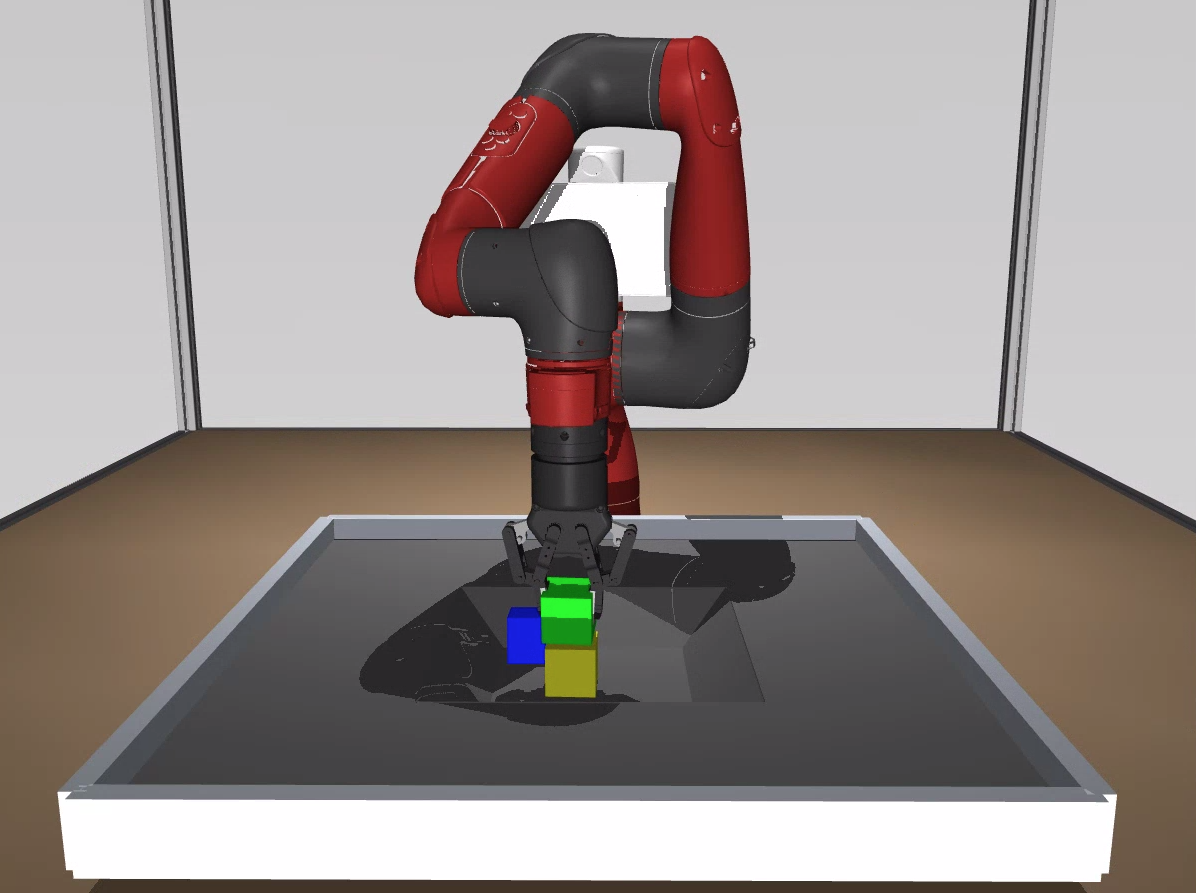}
    \includegraphics[height=0.28\textwidth]{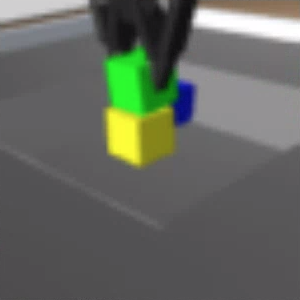} 
    \includegraphics[height=0.28\textwidth]{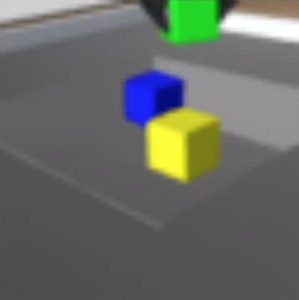} 
    \end{tabular}
    \caption{Left: Overview of the robot manipulation domain. Middle and right: Example robot observations from one camera.
    }
    \label{fig:robot}
    \vspace{-3mm}
\end{minipage}
\end{figure}

\begin{figure*}
\begin{minipage}[b]{\textwidth}
	\centering
	\begin{tabular}{cccc}
         & Lift & Stack & Push \\
        \hspace{-2mm}
        \rotatebox{90}{~~~~~~~Full Model} & \hspace{-3mm}
        \includegraphics[trim=0.3cm 0 0 0, clip,height = 0.17\textwidth]{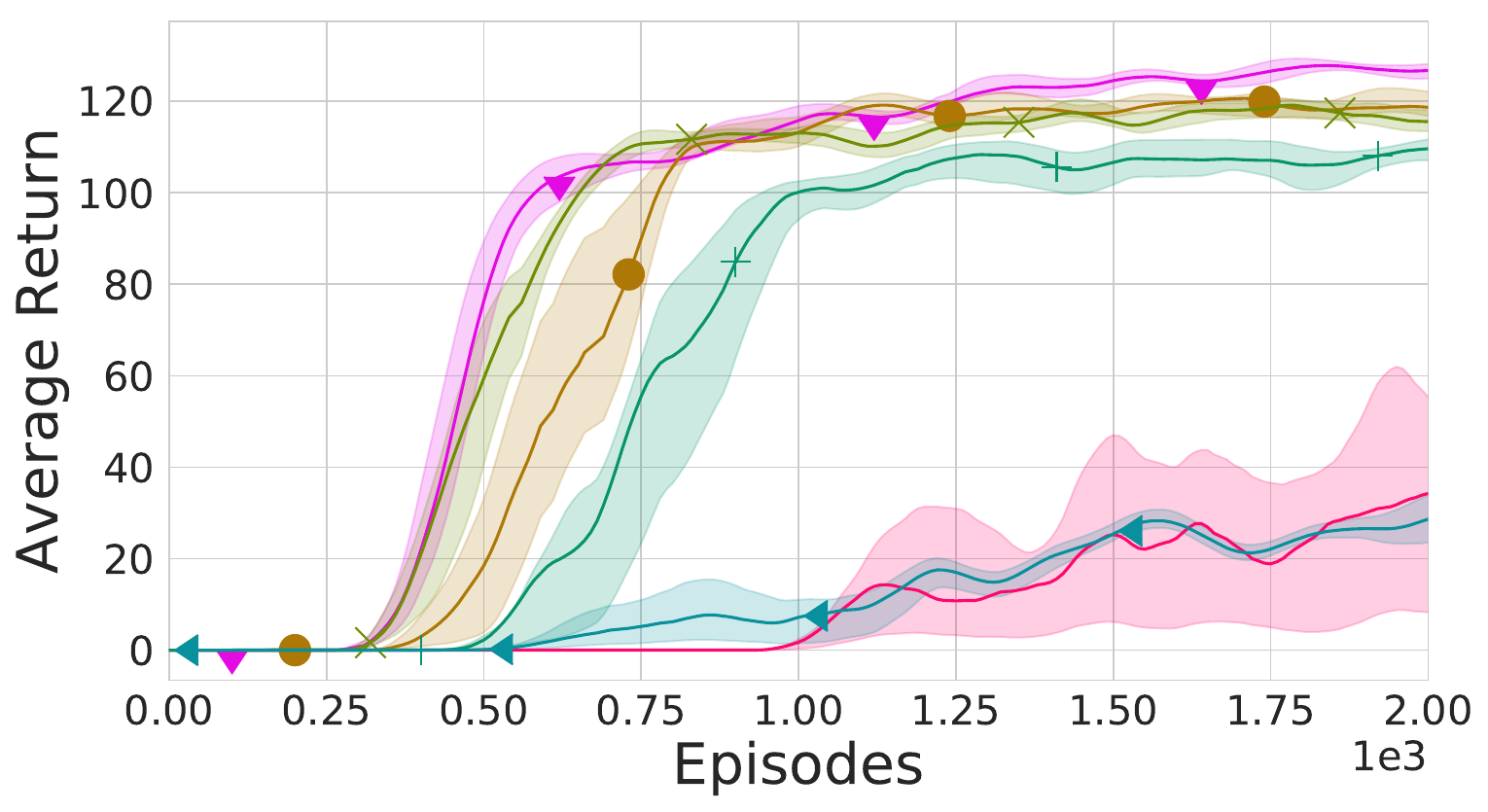} &
        \hspace{-5mm}
        \includegraphics[trim=0.3cm 0 0 0, clip,height = 0.17\textwidth]{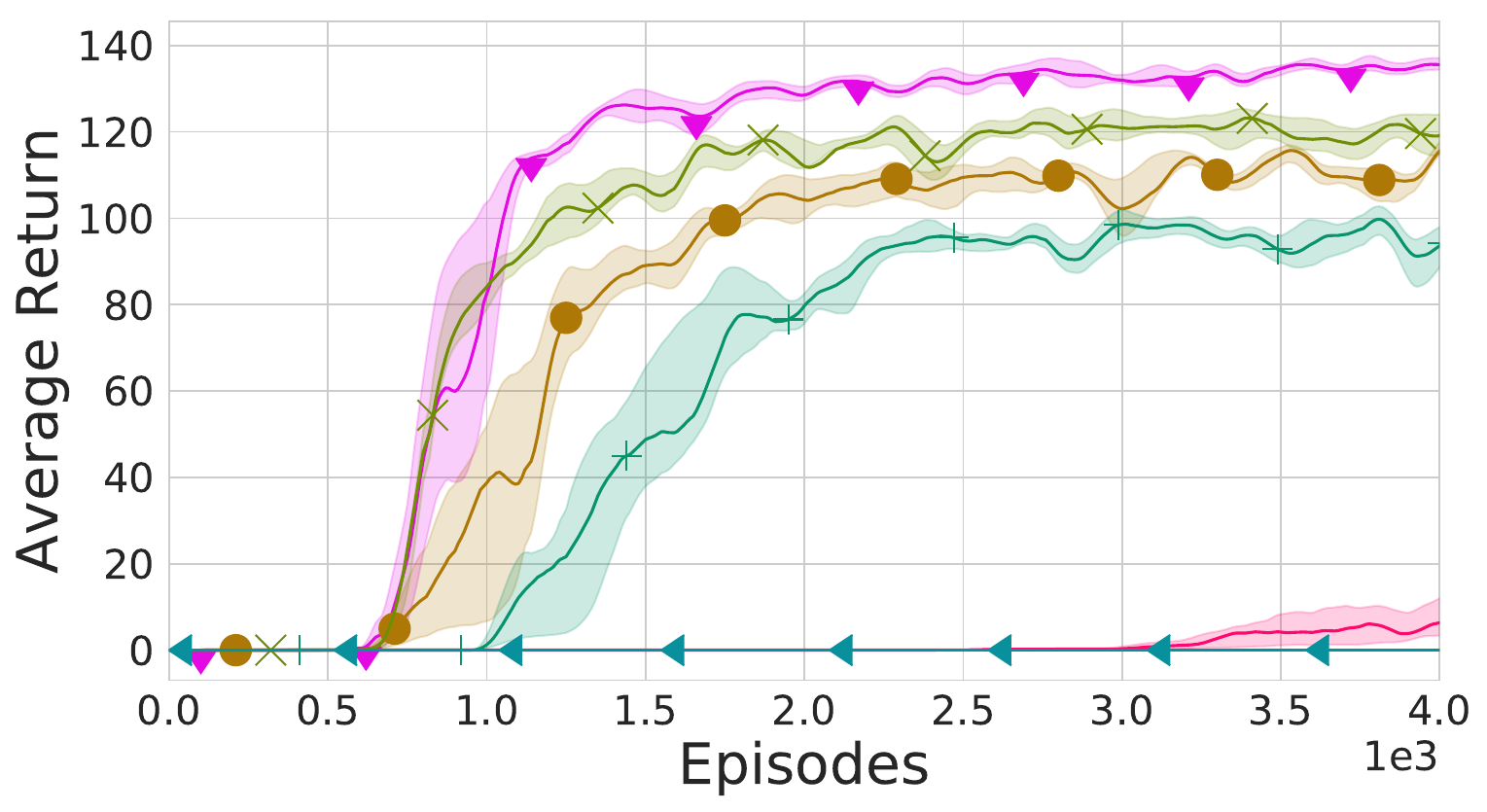} &
        \hspace{-5mm}
        \includegraphics[trim=0.3cm 0 0cm 0, clip,height = 0.17\textwidth]{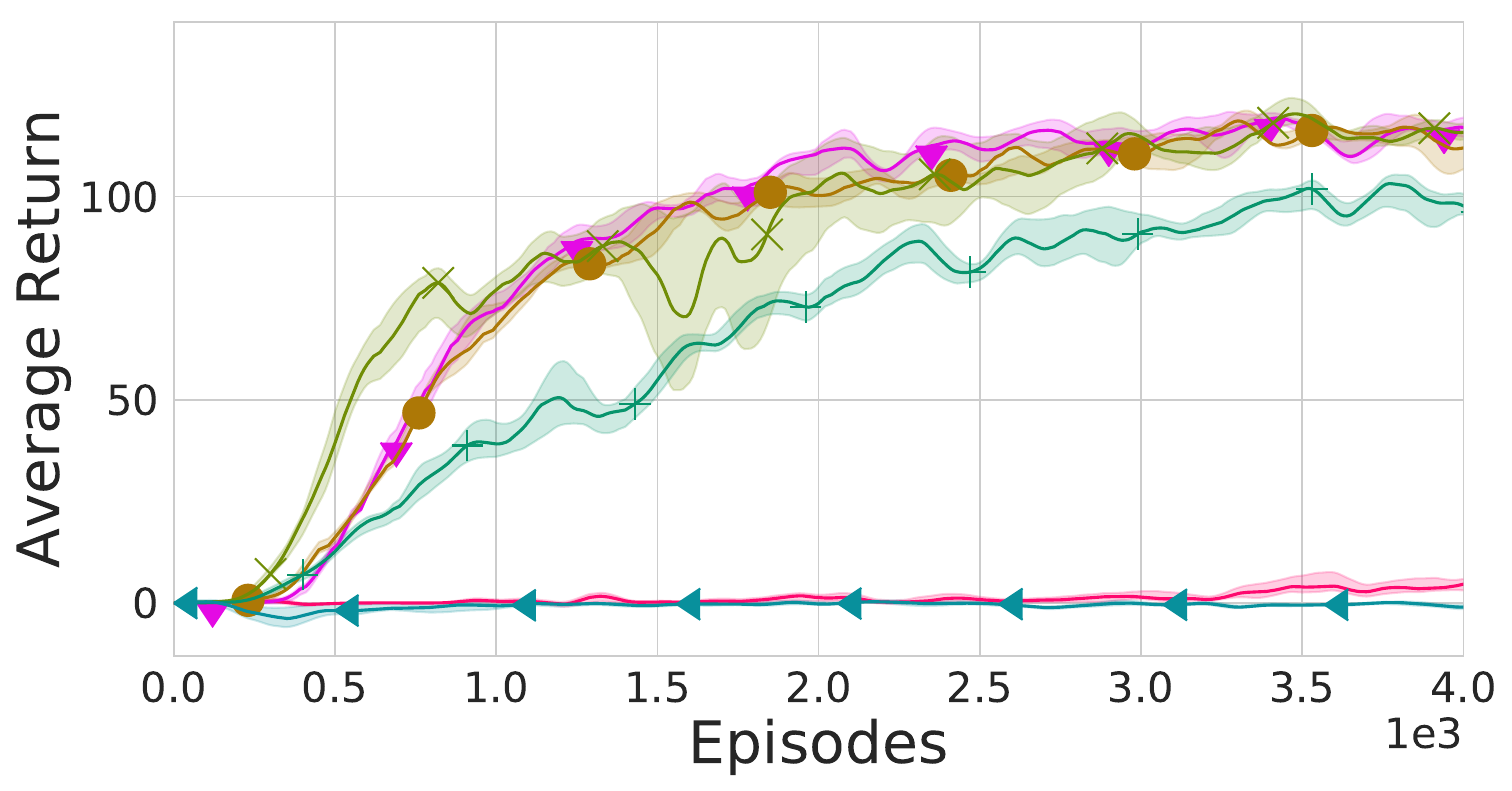}\\
        \hspace{-2mm}
        \rotatebox{90}{~~~~~~Small Model} & \hspace{-3mm}
        \includegraphics[trim=0.3cm 0 0 0, clip,height = 0.17\textwidth]{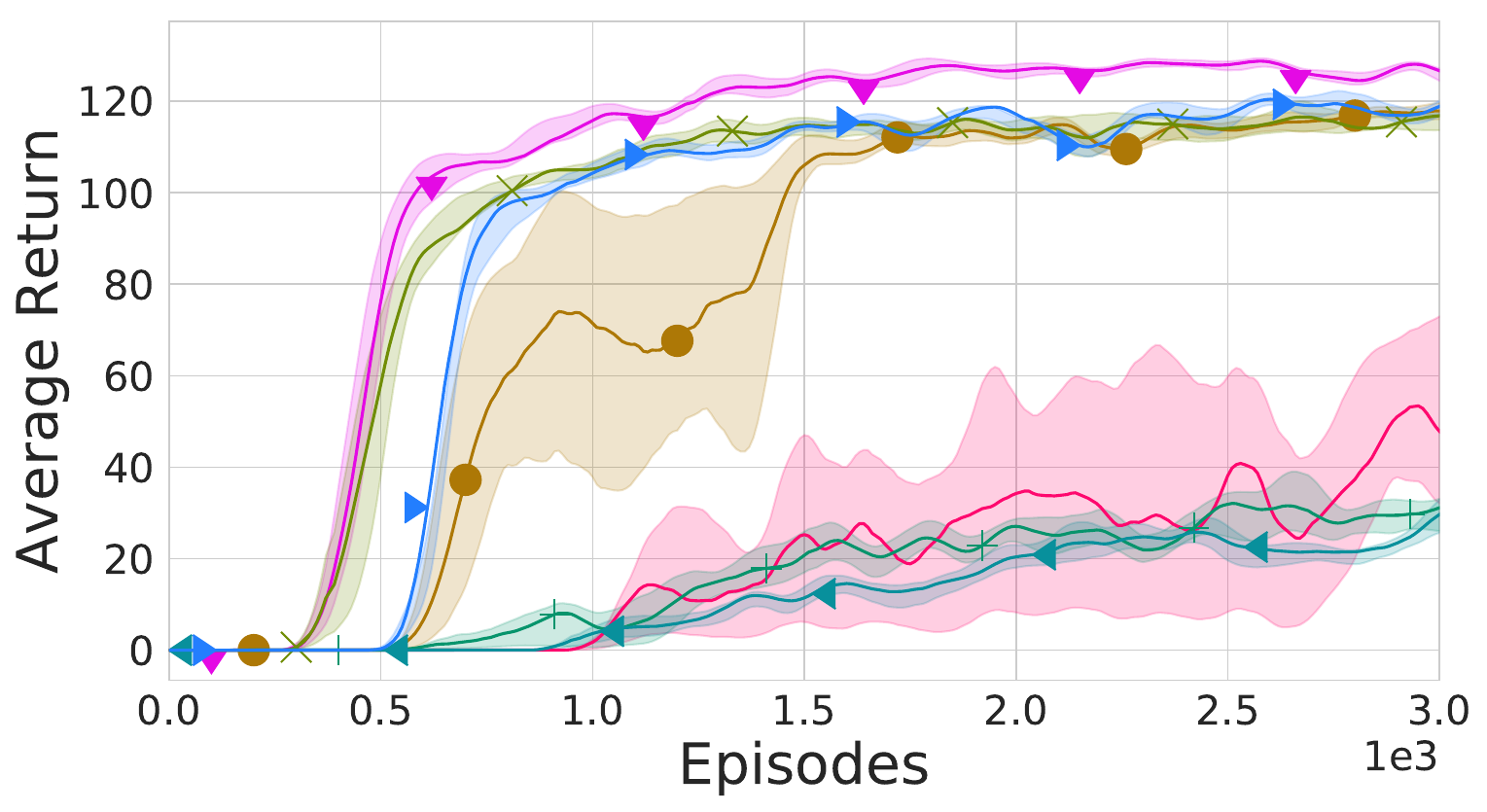} &
        \hspace{-5mm}
        \includegraphics[trim=0.3cm 0 0 0, clip,height = 0.17\textwidth]{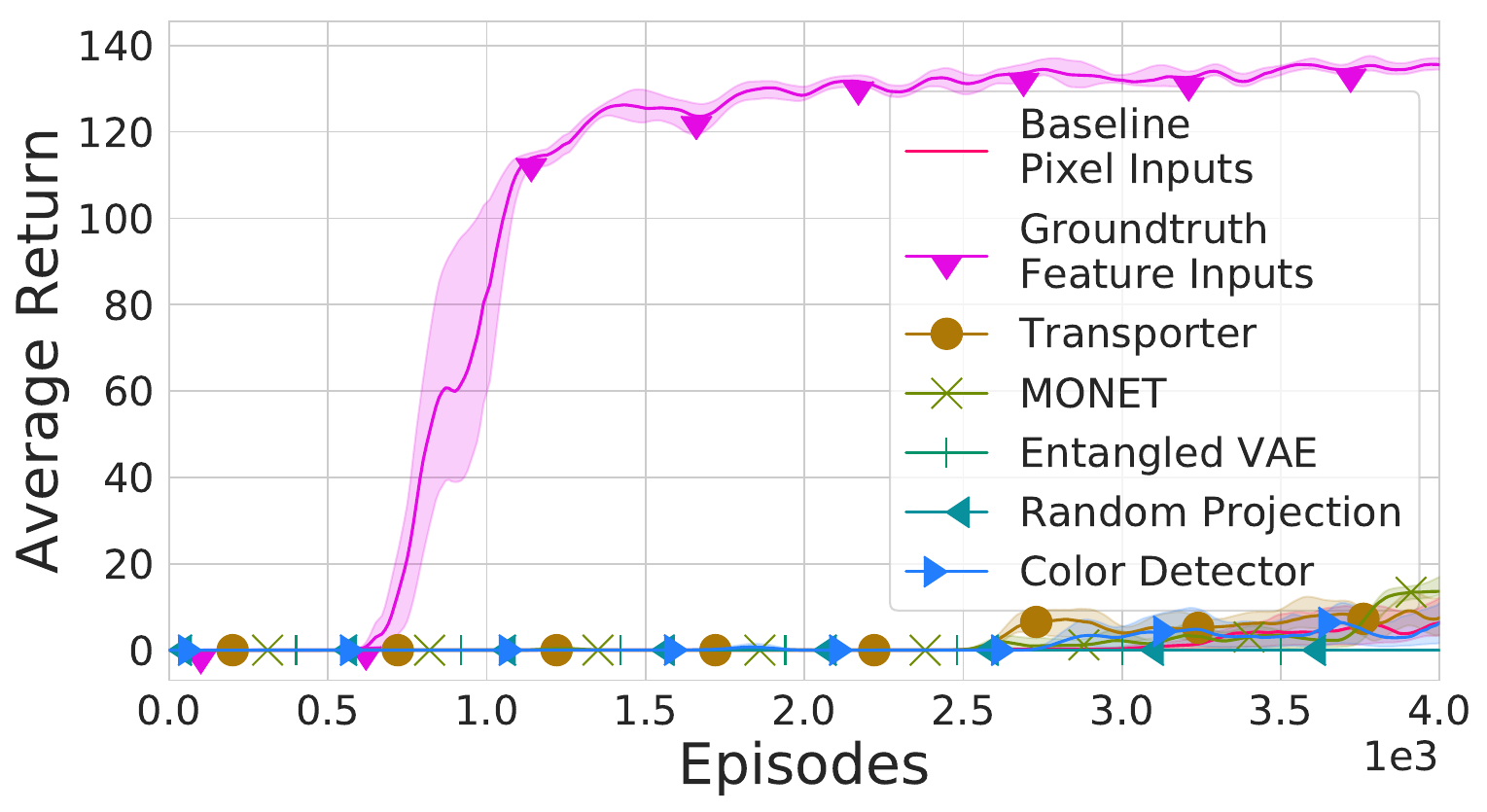}&
        \hspace{-5mm}
        \includegraphics[trim=0.3cm 0 0 0, clip,height = 0.17\textwidth]{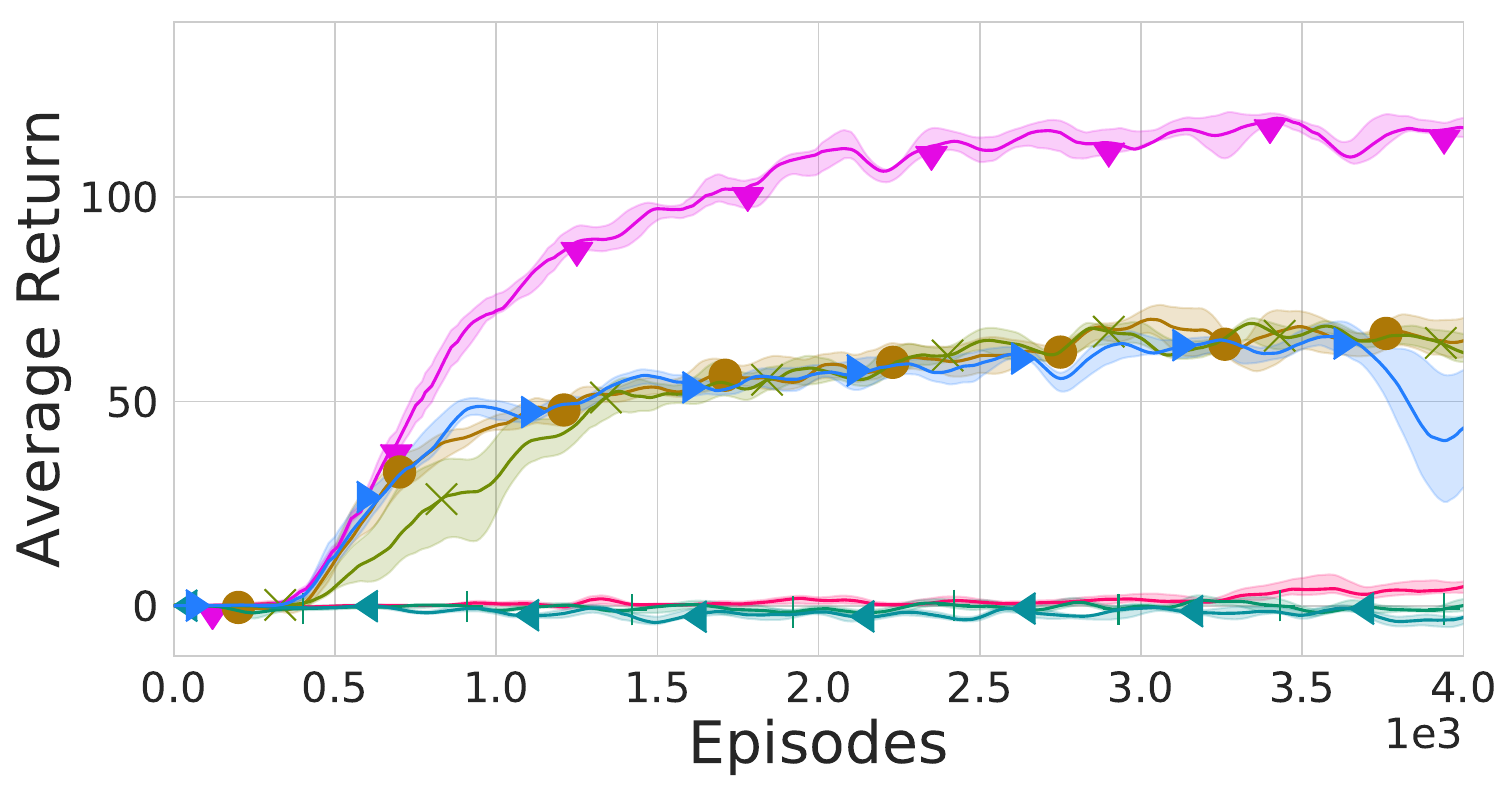}
	\end{tabular}
    \caption{Lifting (left), stacking (middle) and pushing (right) experiments for small (bottom) and full (top) input representations.}
	\label{fig:all_inputs}
	\vspace{-4mm}

\end{minipage}
\end{figure*}

In this section, we perform a set of experiments to investigate how the learned and hand-designed representations described in the previous section can be used as part of an RL agent. Our main focus lies in two applications affecting an agent's perception and exploration, respectively: transforming observations and generating auxiliary tasks (see Figure \ref{fig:main}).

All experiments use a simulated 3D manipulation environment that involves a Sawyer robotic arm interacting with objects on a tabletop (Figure \ref{fig:robot}). We consider three task domains - lifting, stacking and pushing to match object positions - each one requiring successful sequential execution of a number of sub-tasks provided to the agent as hand-engineered auxiliary tasks in the representations-as-inputs scenario. In the lifting domain, the sub-tasks are to reach, grasp and lift the green cube. In the stacking domain, the sequence of tasks from the lift domain is extended to stack the green cube on top of the yellow cube. Finally, position matching requires the arm to push two blocks to specified positions and, so an additional reach auxiliary task is provided for each block.

The final tasks in all domains have sparse rewards and hence require a curriculum of auxiliary tasks to enable efficient exploration \cite{wulfmeier2019regularized}. For example, successful execution of lifting requires a curriculum of first learning how to reach and grasp based on hand-engineered auxiliary tasks. Hence, for the task generation experiments, we remove this handcrafted curriculum to evaluate whether learning to control individual representation dimensions provides sufficient exploration for solving the final task. 

In all experiments, the agent is provided with the proprioception information (e.g. joint angles), and camera images from a third-person viewpoint (64x64 pixels). Depending on the experiment, this image can either be transformed into one of the low-dimensional representations in Sec.~\ref{sec:rep} or fed directly as input to the agent. The latter is also used as a baseline to lower bound the performance in our experiments. 
We pre-train all the representations offline using a dataset that was generated with an agent that was trained on the stacking task; these representation learning models are held fixed (i.e. no fine-tuning) in all our experiments. All models are evaluated in terms of their success on the final task (lifting, stacking or final matched position).

Across the three task domains, and the two use-cases, we evaluate the importance of the different qualities of the learnt representations: \emph{dimensionality}, \emph{observability} and level of \emph{disentanglement}. We follow the symmetry-based definition of \emph{disentanglement} from \cite{higgins2018towards}, which intuitively suggests that a representation is disentangled if it can be decomposed into semantically meaningful and independently transformable sub-spaces. 
For example, if manipulations of single dimensions in a representation result in changes of independent, interpretable factors, like object position, or colour, such a representation is considered disentangled. 
Object-based representations emerge from such a definition, since individual properties of each object can be transformed independently of each other (e.g. changing the colour of one object does not affect the shape of that object or the shape or colour of any other object). 
According to this definition, representations obtained from MONet, Transporter and manual color segmentation can all be considered to be disentangled, while representations from the other methods (VAE, \& Random Projection) are not. 
In terms of \emph{dimensionality}, we use three different variations of each model representation: a) Small (2 dimensional embedding), b) Medium (16 dimensional embedding) and c) Full (size dependent on the model, up to about 100 dims). For methods that learn interpretable, disentangled representations, we train a single model and select appropriately sized subsets of the learnt representation features based on their semantic relevance to the tasks. For the entangled representations where meaningful sub-selection is not possible, we train a separate model for each variant. In terms of \emph{observability}, the Medium and Small conditions may lead to a loss of information compared to the Full condition and hence may lead to partial observability of the state. 
We provide further details on the models and experimental setup in the appendix.

\subsection{Observation Transformation} \label{sec:input_exps}

In this section, we look into using different representations to transform agent inputs from high-dimensional pixel observations to low-dimensional state. Results are presented in Figure \ref{fig:all_inputs}. 
Across all tasks, we compare the performance of different representations to two baselines. 
The pixel baseline, where state representations are learnt implicitly from pixels via end-to-end reinforcement learning, serves as the lower bound. This baseline fails to achieve good performance in all tasks in the given training time.
The upper bound baseline is provided by the simulator features used as the low-dimensional, structured input (about 100 dimensions). Note that both baselines use the full observations in all scenarios (Full, Medium and Small). Impressively, MONET, Transporter and VAE often obtain performance close to the feature baseline, for both the medium and full model of the representation, suggesting that the exact size of representation ($\ll100$ dimensions in our case) has only limited effect as long as the environment remains fully observable via the representation. For the bigger models, we also find limited difference in performance between disentangled representations, such as MONET and Transporter, and entangled ones, such as the entangled $\beta-$VAE, across both tasks. 
There is, however, a potential engineering benefit when using smaller representations. Here, disentangled representations enable us to successfully choose smaller subsets of the full representations while ensuring that the information relevant for the task is not discarded. Indeed, all the disentangled models which enable us to include all the relevant information for lifting (MONet, Transporter, Color Detector) solve the task fairly quickly, while the entangled models fail. This highlights the practical importance of the human semantic interpretability of disentangled representations. 
Discarding task-relevant dimensions in disentangled representations unsurprisingly does significantly reduce task performance. This effect becomes visible for the smallest model class (with 2 disentangled dimensions) for the stacking and pushing tasks, where at least 4 features are necessary to represent the positions of the 2 objects to successfully solve the task. Smaller models, which represent the whole observation in two entangled dimensions (VAE, and random projection), should in theory be able to solve the Stack task, since the two dimensions could still contain all the information from the original scene, albeit projected into a 2D embedding. This is definitely the case for the VAE, since the embedding is sufficient to reconstruct the original image. However, these models fail to solve the task, suggesting the importance of the representational form of the embedding.

\subsection{Auxiliary Tasks} \label{sec:output_exps}
\begin{figure}[h]
\begin{minipage}[b]{\linewidth}
	\centering
	\begin{tabular}{cc}
        Medium Model & Small Model\\
        \includegraphics[trim=0 0 0 0, clip,height = 0.3\textwidth]{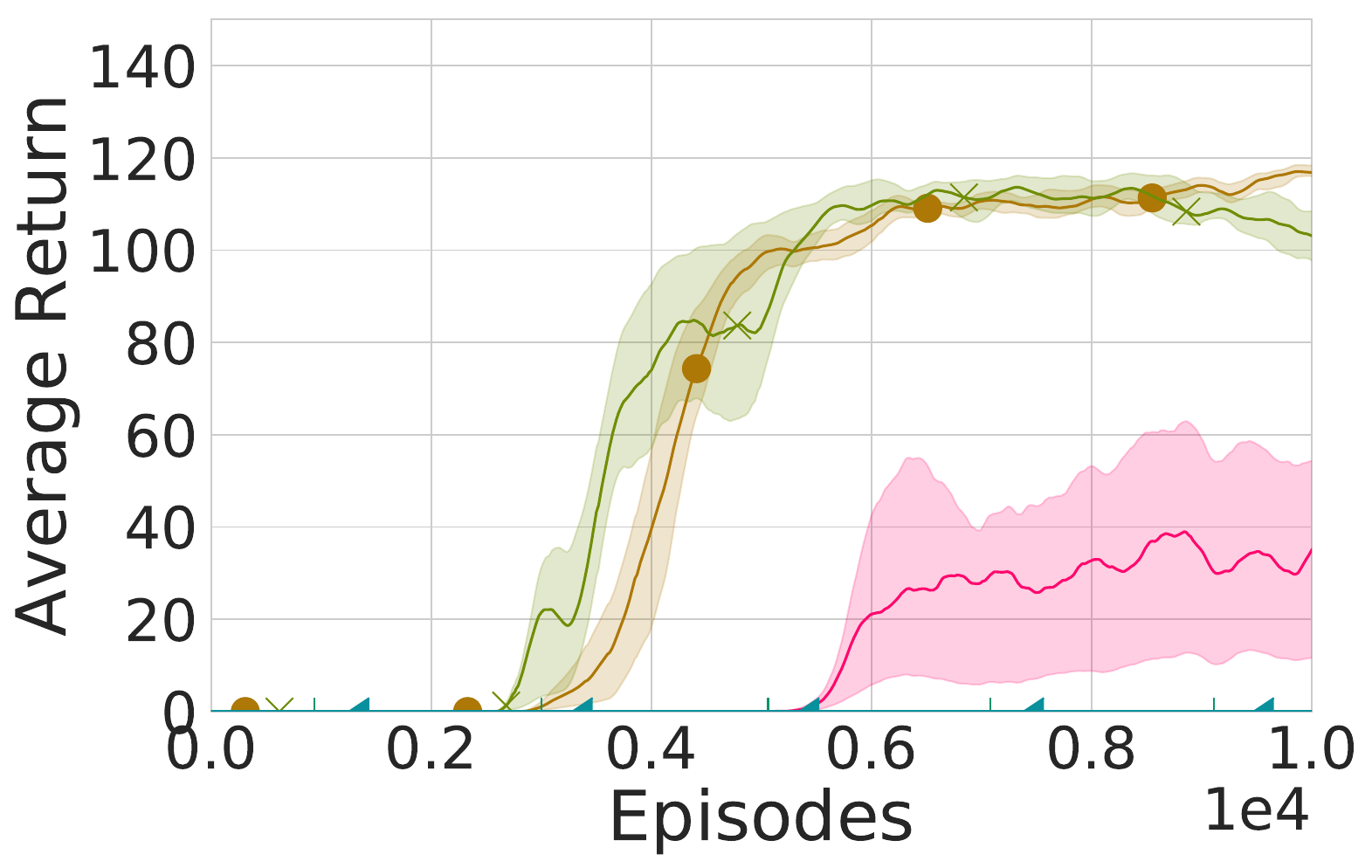} &
        \hspace{-4mm}
        \includegraphics[trim=0 0 0cm 0, clip,height = 0.3\textwidth]{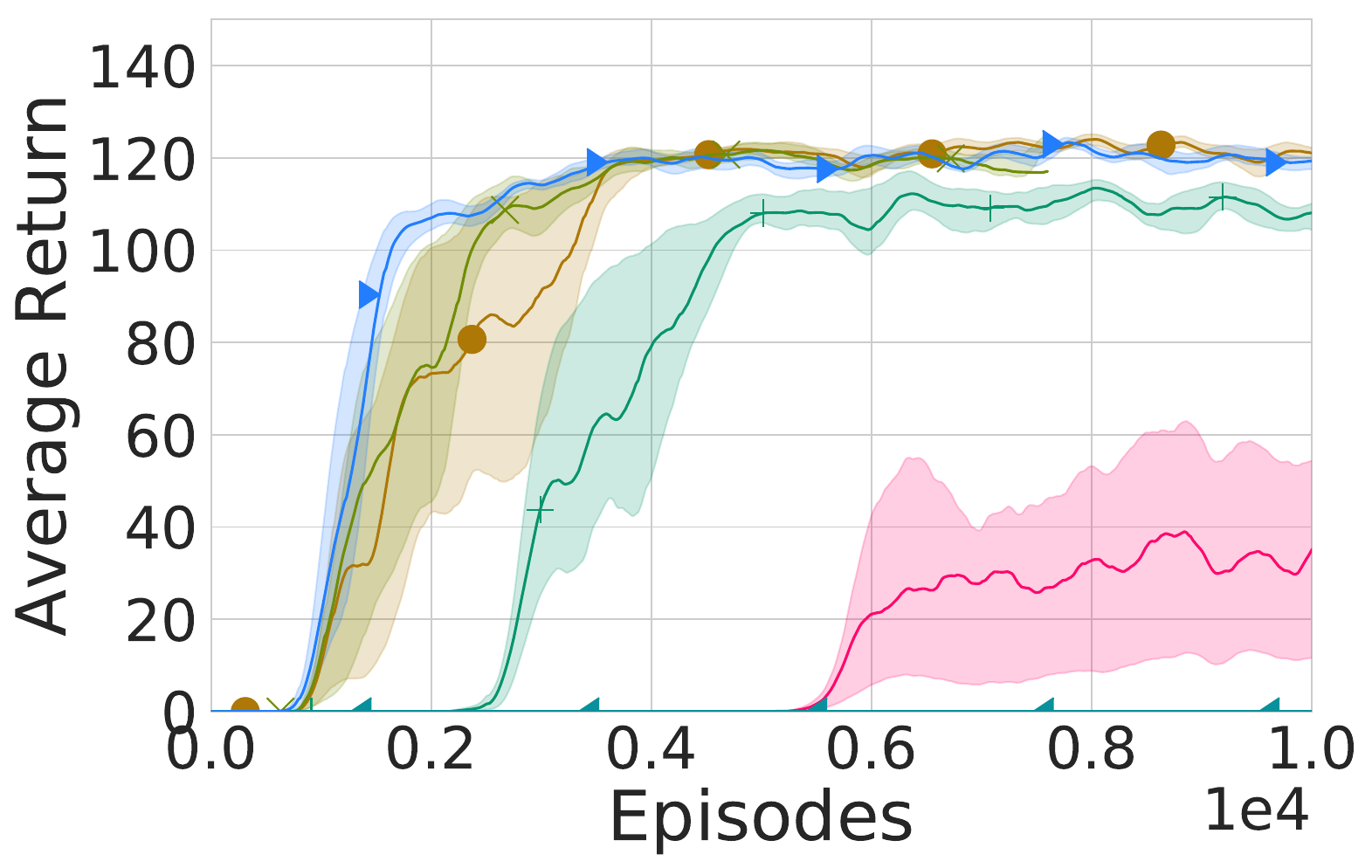}\\
        \includegraphics[trim=0 0 0 0, clip, height = 0.3\textwidth]{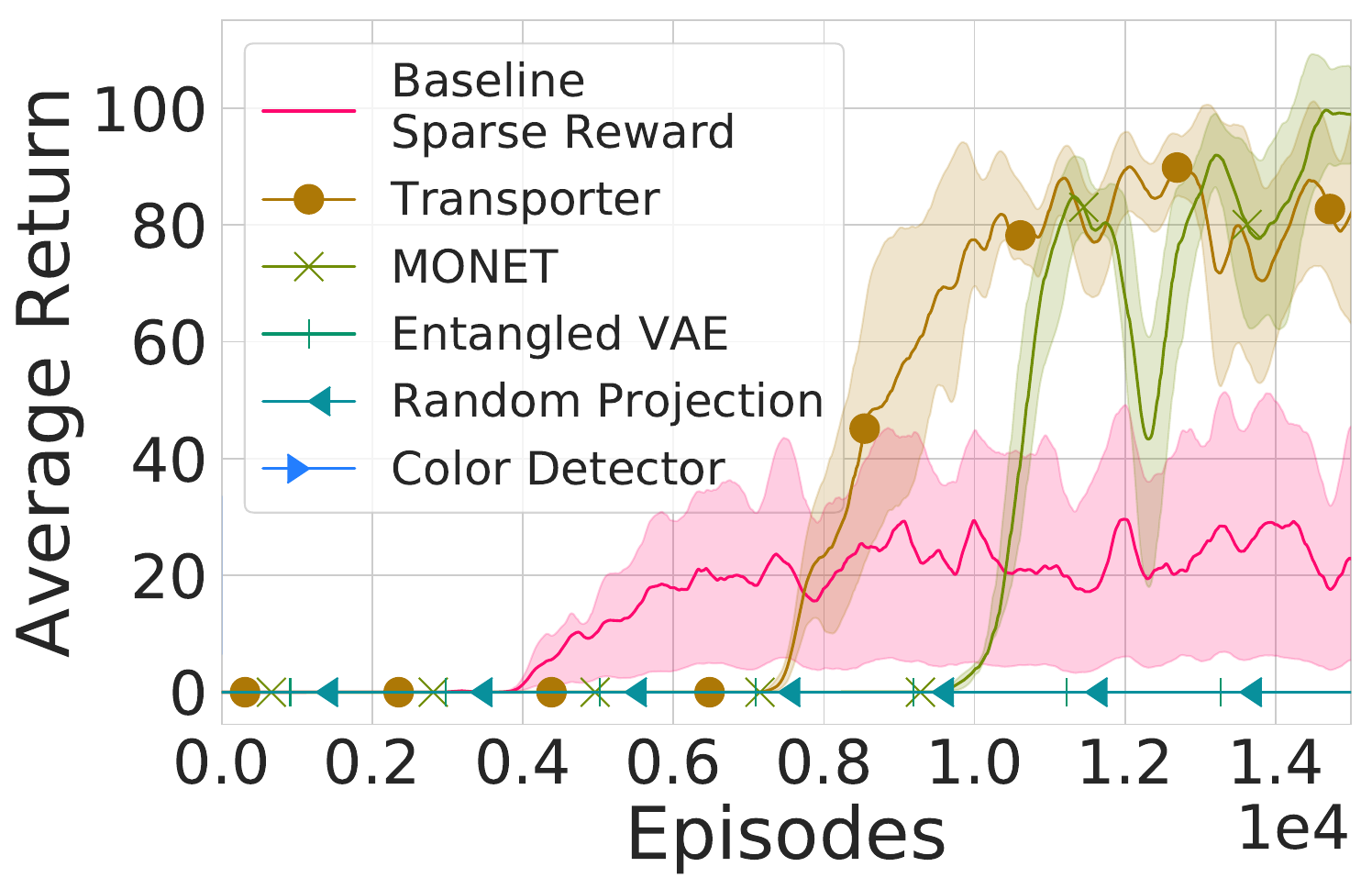} &
        \hspace{-4mm}
        \includegraphics[trim=0 0 0 0, clip, height = 0.3\textwidth]{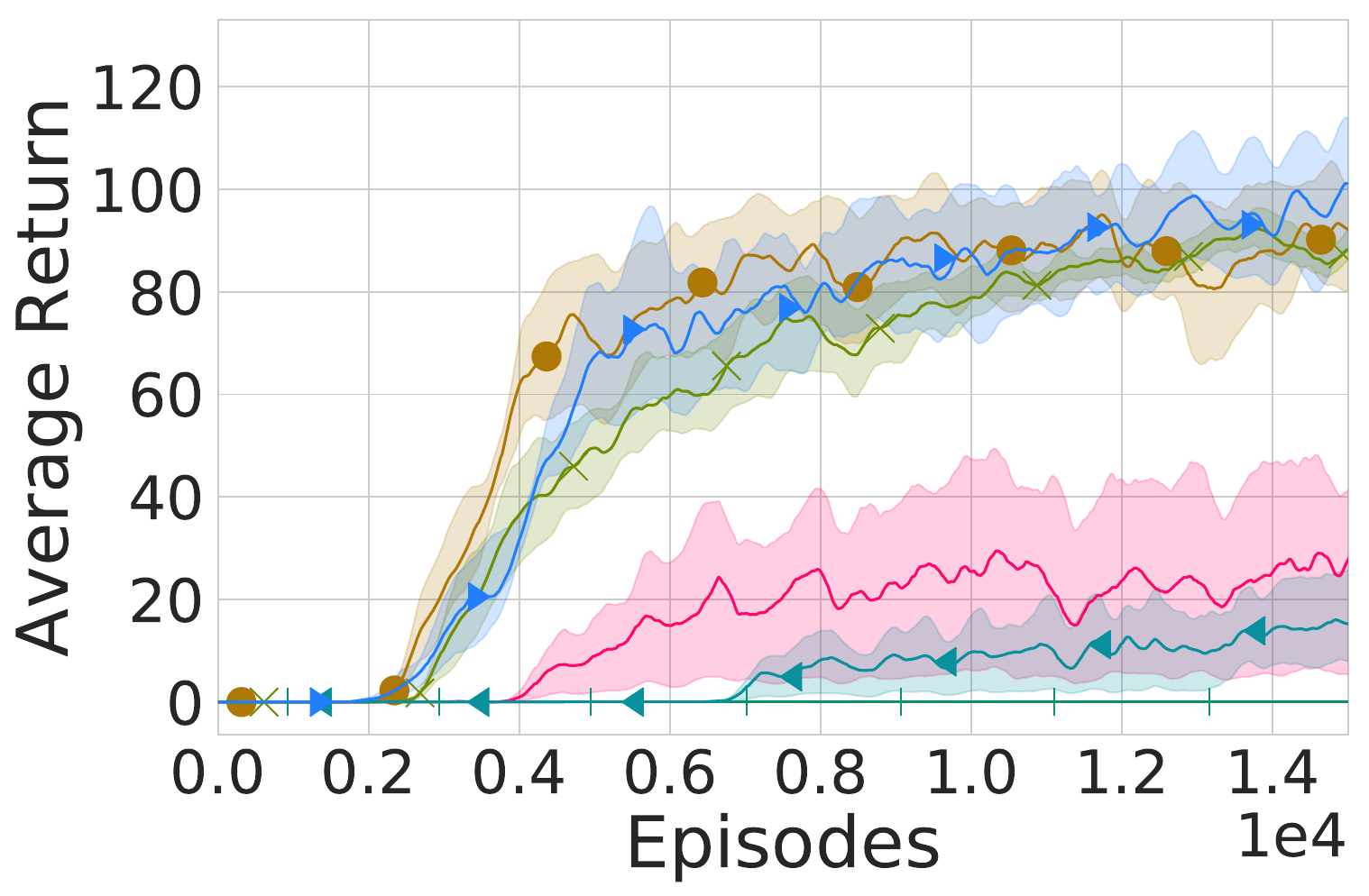} \\
        \includegraphics[trim=0 0 0 0, clip, height = 0.3\textwidth]{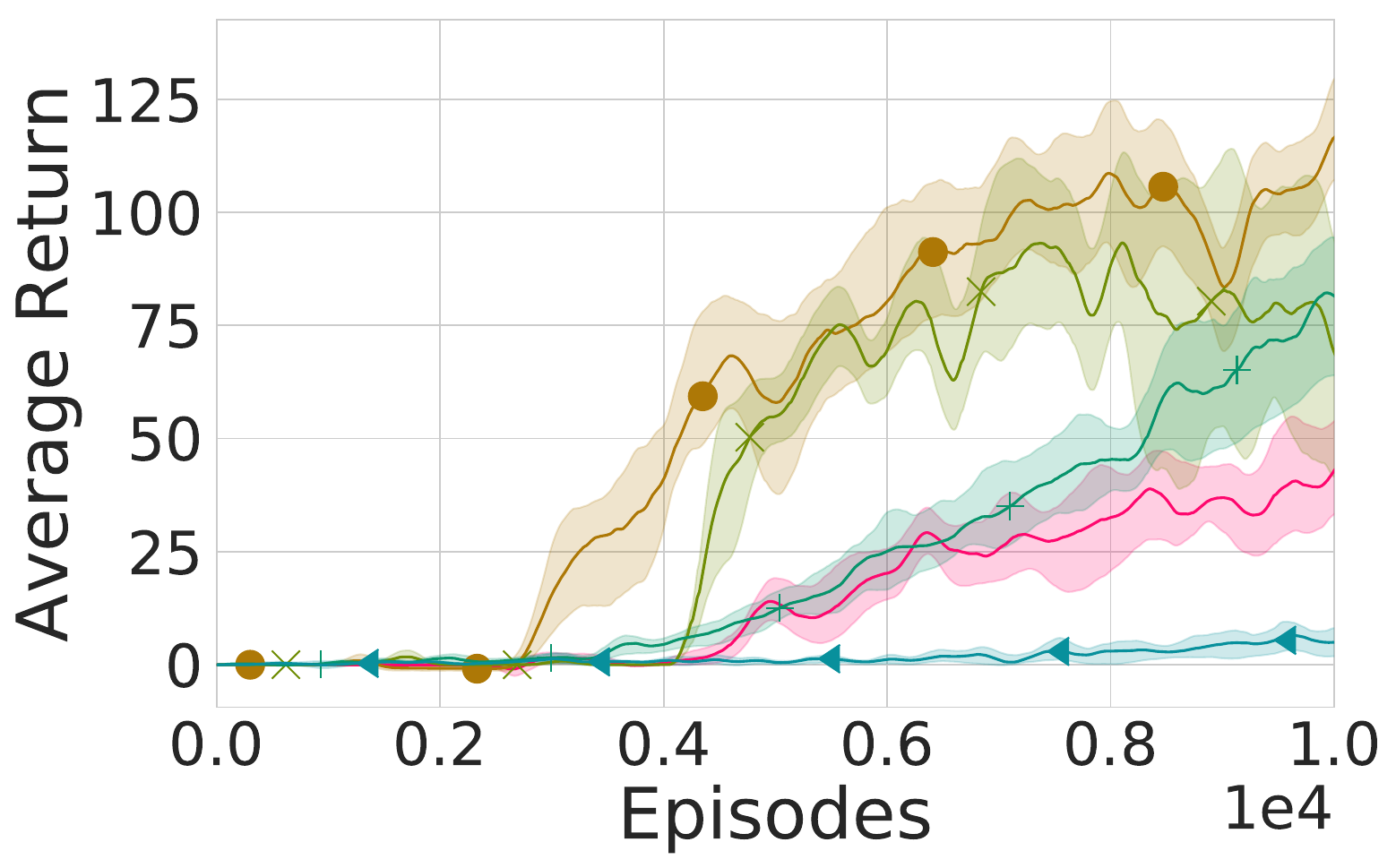} &
        \hspace{-4mm}
        \includegraphics[trim=0 0 0 0, clip, height = 0.3\textwidth]{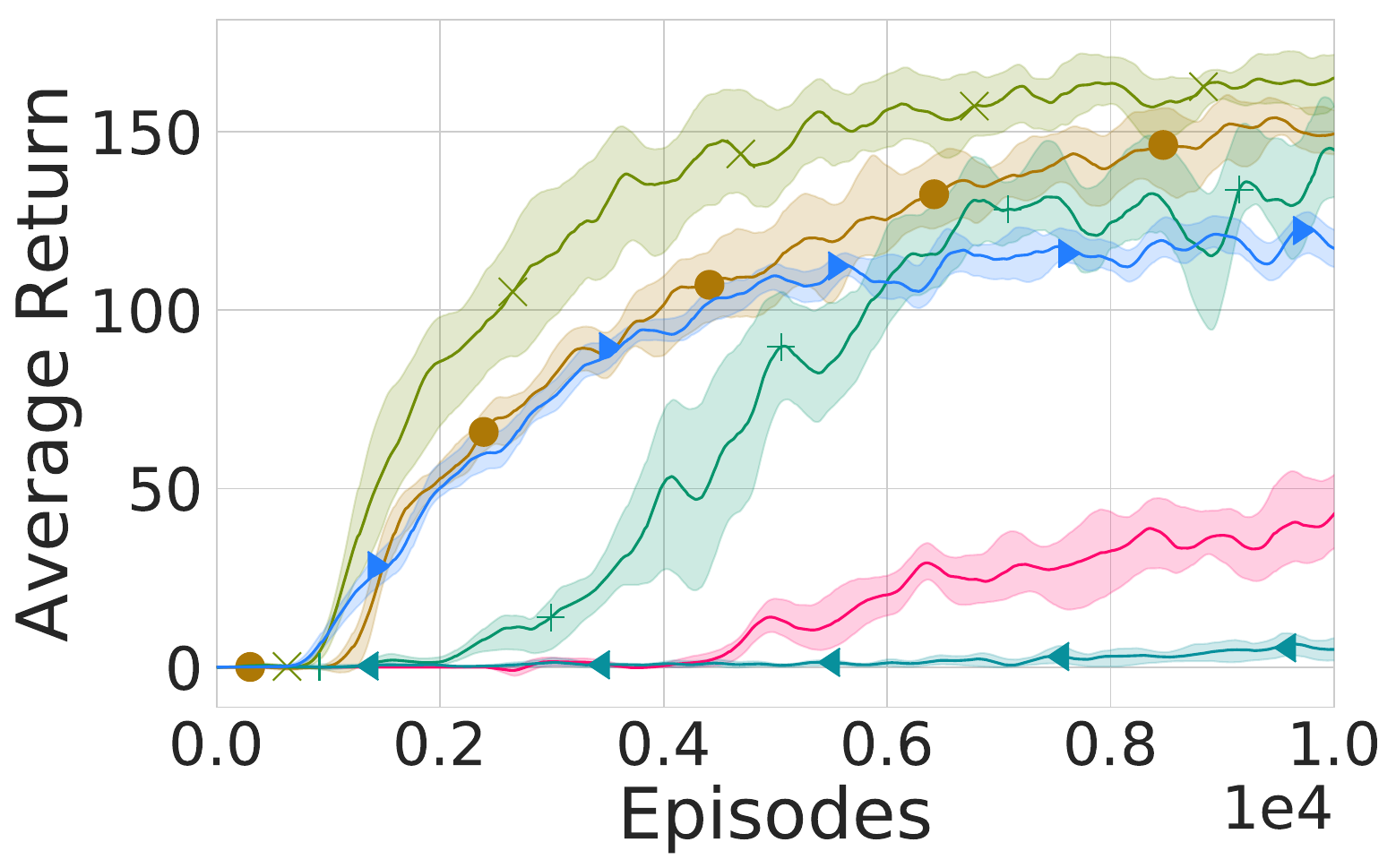}         
	\end{tabular}
    \caption{Lifting, stacking \& pushing (top/middle/bottom) experiments for small \& medium task sets with all representations. 
    }
	\label{fig:all_tasks}
	\vspace{-4mm}
\end{minipage}
\end{figure}

To investigate the role of using representations as a source of auxiliary tasks, we use feature observations as inputs to all the models, but replace the hand-engineered auxiliary tasks used in the previous section with the representation feature control tasks from Section \ref{sec:tasks}. We focus on the Small model (4 tasks from 2 dimensions) and Medium model (32 tasks from 16 dimensions) conditions from Section \ref{sec:input_exps}.  
The largest Full model representations have been excluded from this evaluation due to computational constraints.
In this scenario all generated data is shared across tasks, which in practice means that each transition gets assigned rewards for all auxiliary tasks and the final task. Given that we use learnt representations as image transformation, we can assign auxiliary task rewards to stored data on the fly during learning.
Results are presented in Figures \ref{fig:all_tasks}.

Even though the tasks are defined by sparse rewards, we see that some individual seeds for the pixel baseline with no auxiliary tasks can get lucky and gain enough signal to achieve reasonable performance. The majority of the seeds, however, never get any reward. Unlike the results presented in the previous section, when it comes to using representation features for auxiliary tasks, representation dimensionality does matter. Higher dimensional representations create additional learning challenges, requiring to generate data on more tasks and to represent the solution to bigger task sets \cite{yu2020meta, wulfmeier2019regularized}. 
This results in slower learning of the final task for Medium vs Small models. Furthermore, we see that disentanglement matters more in this scenario. Tasks based on MONet, Transporter (and color detection for the small model) enable the agent to learn how to manipulate independently controllable aspects of the environment \cite{grimm2019disentangled, laversannefinot2018curiosity}. Entangled models, such as the entangled $\beta$-VAE, do not perform well in comparison. Intuitively, learning a policy to control individual dimensions of an entangled representation is hard. Note that partial observability of the state (Small Model) is not as much of an issue in the auxiliary-task scenario compared to the representations-as-inputs scenario. This is because learning how to control even a small subset of task-relevant features is enough to help bootstrap final task learning. For more complex and sparser tasks however, partial observability could become a challenge also for task generation.

\subsubsection{Bigger Task Sets}\label{sec:scheduling_exps}
\begin{figure}[t]
	\centering
	\begin{tabular}{c c}
        \includegraphics[trim=0 0 6.6cm 0, clip,width =  0.26\textwidth]{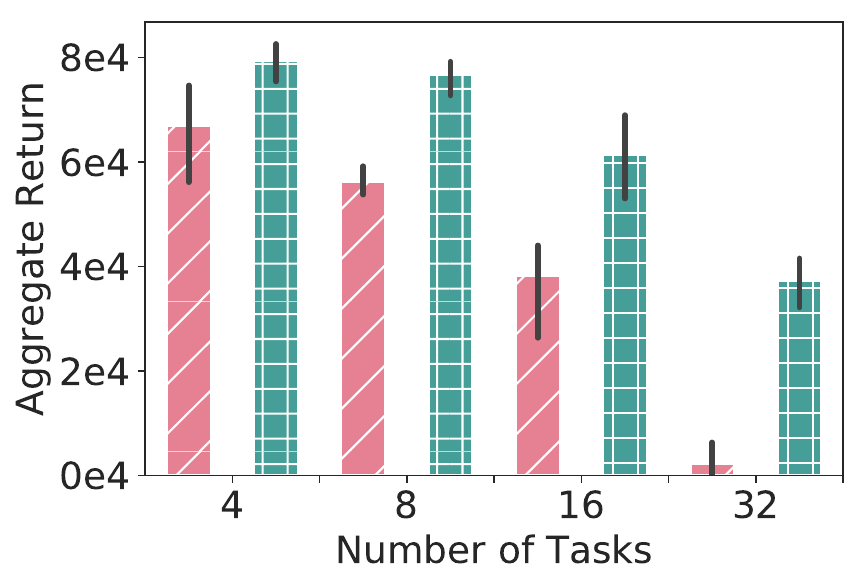}
        \includegraphics[trim= 14.4cm 0 0 0, clip,width =  0.13\textwidth]{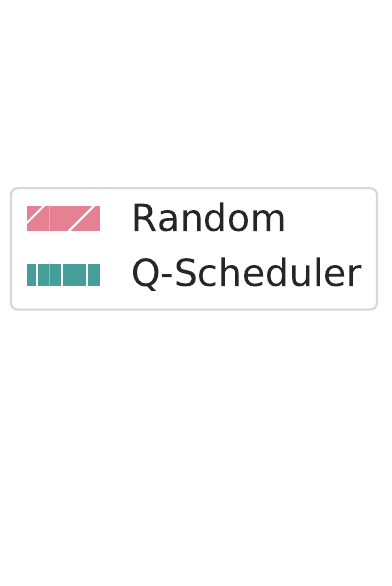}
	\end{tabular}
    \caption{Scheduling ablations for increasing number of tasks. Left: Learning curves; Right: Aggregated return across learning. We use 4 to 32 tasks from the representations plus the main task. Increasing the number of tasks reduces performance and increases data requirements and scheduling beneficial auxiliary tasks becomes more important.}
	\label{fig:scheduling}
	\vspace{-4mm}
\end{figure}

Increasing the number of tasks generally leads to an increase in the training time of any agent to complete the final task. The bigger the space of tasks, the more important becomes the choice of which which tasks to learn from.
In this section we compare randomly scheduling executed tasks against using the Q-Scheduler approach detailed in Section \ref{sec:scheduling}, which optimizes the task choice to maximize performance on the final task. Figure \ref{fig:scheduling} visualizes agent performance and demonstrates that  Q-scheduler improvements become more emphasized for bigger task sets.
\section{Related Work}\label{sec:related}

\textbf{Representation Learning:} Learning representations from high-dimensional observations such as images can considerably increase data-efficiency and robustness for down-stream tasks.
Reducing the human effort for data-collection and labelling, unsupervised and self-supervised objectives, produce benefits for training compressed representations e.g. via reconstruction, beginning from \cite{hinton2006reducing}. 
Variational Auto-Encoders (VAEs) combined this direction with techniques from variational inference \cite{kingma2013auto, rezende2014stochastic}.
Several approaches extend VAEs towards learning disentangled object-centric representations of scenes, notably $\beta$-VAEs \cite{higgins2017beta} and MONet \cite{burgess2019monet}. As an alternative to reconstruction, several other self-supervised approaches have been explored for representation learning, including through the use of predictive \cite{kulkarni2019unsupervised}, discriminative \cite{donahue2019large} and contrastive \cite{oord2018representation, chen2020simple, grill2020bootstrap} losses.

\textbf{Representations as Agent Inputs:} Learning policies and value functions directly from raw visual observations has been successful for several tasks both in simulation \cite{mnih2013playing, lillicrap2015continuous} and in the real world \cite{riedmiller2018learning, kalashnikov2018qt}. 
While these successes are encouraging, learning from pixels is commonly less sample efficient and robust than from hand-crafted simulator features \cite{tassa2018deepmind, kaiser2019model}. 
To alleviate this, several recent approaches learn intermediate feature representations through the use of auxiliary losses such as reconstruction \cite{byravan2020imagined, hafner2019dream, lee2019stochastic,locatello2019challenging}, 
contrastive losses \cite{oord2018representation, srinivas2020curl, grill2020bootstrap} and future prediction \cite{schwarzer2020data, kulkarni2019unsupervised}.
Unlike many of these methods which learn feature representations as part of the RL learning loop, a few approaches explored the use of offline learned features for reinforcement learning \cite{lange2012autonomous, aytar2018playing, higgins2017darla, laversannefinot2018curiosity, nair2018visual}. In this paper, we take a similar line and pre-train several state of the art representation learning models and evaluate their efficacy on multiple simulated manipulation tasks.

\textbf{Representations for Auxiliary Tasks:} Auxiliary tasks have been used in reinforcement learning to generate additional training regularization signals 
\cite{Dosovitskiy2017,Jaderberg2017Unreal,Mirowski16,cabi2017,kulkarni2019unsupervised,higgins2017darla} as well as to improve exploration \cite{kaelbling1993learning,riedmiller2018learning, wulfmeier2019regularized, hertweck2020simple, andrychowicz2017hindsight, wulfmeier2020data}. 
In both cases, the observed benefits depend on the nature and difficulty of the auxiliary tasks, and how these implicitly affect the nature of the learnt representation within the RL agent. However, no principled recipe currently exists for choosing the right set of such tasks.  
Our work is extending the existing line of work that tries to get a better understanding of what representation properties are desirable and why, before using this knowledge to improve the existing methods \cite{grimm2019disentangled,sharma2019dynamics,riedmiller2018learning,wulfmeier2019regularized, hertweck2020simple,relevantfreatures20}.
Although our work touches on exploration, it is different and this perspective is complimentary to the previous work that
improved exploration via diversity \cite{sharma2019dynamics, eysenbach2018diversity}, 
curiosity \cite{pathak2017curiosity,Oudeyer17},
or empowerment objectives \cite{klyubin2005all, gregor2016variational}.
Indeed, many of these methods can benefit from insights into representation learning and  
recent work has shown promising results for manually-chosen features to guide skill learning \cite{sharma2019dynamics,riedmiller2018learning,wulfmeier2019regularized, hertweck2020simple} as well as features from an agent's value function \cite{relevantfreatures20}.
Finally, choosing which task to execute given a large set of options creates an additional exploration-like challenge. 
Random sampling often provides a strong baseline \cite{graves2017automated} but can be improved via curriculum learning \cite{bengio2009curriculum, heess2017emergence, Oudeyer17,florensa2017reverse} as well as iterative generation of tasks \cite{Schmidhuber2013PowerPlayTA,wang2019paired}. Here, we extend the branch of work on task scheduling \cite{riedmiller2018learning}.


\section{Discussion and Conclusions} 
\label{sec:analysis}
In this paper, we systematically evaluate how representations with different properties can help learning on downstream tasks in three challenging robotics settings. We are able to considerably improve performance and data-efficiency. For example, we show that MONet and Transporter representations can perform on par with simulator states when used as agent inputs. 
We investigate the benefits of representations principally along three dimensions: dimensionality, partial observability and disentanglement; and using two use-cases: representations-as-inputs and representations for auxiliary tasks. While the three dimensions used to compare our representations are not sufficient to detail all differences between the models, they provide useful intuitions on which representations may perform well and why. 

\textbf{Dimensionality} has considerable impact for task generation. More tasks lead to higher data requirements even if we assign rewards from all tasks to data independent of the generating task. 
Optimizing which tasks, and related policies, to use to generate data can partially mitigate this effect. For input representations, 
performance remains mostly stable. 

\textbf{Partial observability} has stronger negative effect when representations are used as inputs rather than sources of auxiliary tasks. 
For example, in the stacking experiments, the lack of information about the location of the lower block in the representation makes the task unsolvable in the former use-case, but not the latter. 
When generating auxiliary tasks, partial observability means that exploration with respect to the non-observable factors is lacking. However, depending on the domain, positive exploration gains can be obtained from controlling even a small observable subset of factors.

\textbf{Disentanglement} has limited impact for policy inputs in our experiments, with entangled and disentangled models performing near equivalently for most tasks. However, when forced to use smaller models, the interpretability provided by disentanglement makes it possible to choose a subset of dimensions that are the most important for solving the task, as was demonstrated for the lift experiments. The biggest benefit of disentanglement is observed in the auxiliary task experiments. Learning how to control individual disentangled dimensions allows the agent to explore the state space more efficiently.

The current work 
uses a single simulated robotics domain and three downstream tasks. We hope to extend this to a broader set of environments and tasks, including real robot experiments, in future work. Another direction of future work is to extend the set of representation learning methods. 
For example, here we focus on reconstruction-based representations, however recent improvements in contrastive learning open a new class of representation learning approaches for evaluation.
Finally, one of the biggest shortcomings of our work is the pretraining of representations from existing datasets. This requires good coverage of the domain of interest. 
Hence, another pressing direction for future work is the online training of representations from the agent's own experience.

\section*{Acknowledgements}{
The authors would like to thank Yusuf Aytar, Volodymyr Mnih, Nando de Freitas, and Nicolas Heess for helpful discussion and relevant feedback for shaping our submission.
We additionally like to acknowledge the support of the DeepMind robotics lab for infrastructure and engineering support.
}

\bibliographystyle{IEEEtran}
\bibliography{references}

\begin{thebibliography}{10}
\providecommand{\url}[1]{#1}
\csname url@rmstyle\endcsname
\providecommand{\newblock}{\relax}
\providecommand{\bibinfo}[2]{#2}
\providecommand\BIBentrySTDinterwordspacing{\spaceskip=0pt\relax}
\providecommand\BIBentryALTinterwordstretchfactor{4}
\providecommand\BIBentryALTinterwordspacing{\spaceskip=\fontdimen2\font plus
\BIBentryALTinterwordstretchfactor\fontdimen3\font minus
  \fontdimen4\font\relax}
\providecommand\BIBforeignlanguage[2]{{%
\expandafter\ifx\csname l@#1\endcsname\relax
\typeout{** WARNING: IEEEtran.bst: No hyphenation pattern has been}%
\typeout{** loaded for the language `#1'. Using the pattern for}%
\typeout{** the default language instead.}%
\else
\language=\csname l@#1\endcsname
\fi
#2}}

\bibitem{hinton2006reducing}
G.~E. Hinton and R.~R. Salakhutdinov, ``Reducing the dimensionality of data
  with neural networks,'' \emph{Science}, vol. 313, no. 5786, pp. 504--507,
  2006.

\bibitem{kingma2013auto}
D.~P. Kingma and M.~Welling, ``Auto-encoding variational bayes,'' in
  \emph{ICLR}, 2014.

\bibitem{higgins2017beta}
I.~Higgins, L.~Matthey, A.~Pal, C.~Burgess, X.~Glorot, M.~Botvinick,
  S.~Mohamed, and A.~Lerchner, ``Beta-vae: Learning basic visual concepts with
  a constrained variational framework.'' \emph{ICLR}, vol.~2, no.~5, p.~6,
  2017.

\bibitem{gregor2016towards}
K.~Gregor, F.~Besse, D.~J. Rezende, I.~Danihelka, and D.~Wierstra, ``Towards
  conceptual compression,'' in \emph{NeurIPS}, 2016, pp. 3549--3557.

\bibitem{burgess2019monet}
C.~P. Burgess, L.~Matthey, N.~Watters, R.~Kabra, I.~Higgins, M.~Botvinick, and
  A.~Lerchner, ``Monet: Unsupervised scene decomposition and representation,''
  \emph{arXiv preprint arXiv:1901.11390}, 2019.

\bibitem{kulkarni2019unsupervised}
T.~D. Kulkarni, A.~Gupta, C.~Ionescu, S.~Borgeaud, M.~Reynolds, A.~Zisserman,
  and V.~Mnih, ``Unsupervised learning of object keypoints for perception and
  control,'' in \emph{NeurIPS}, 2019, pp. 10\,723--10\,733.

\bibitem{oord2018representation}
A.~v.~d. Oord, Y.~Li, and O.~Vinyals, ``Representation learning with
  contrastive predictive coding,'' \emph{arXiv preprint arXiv:1807.03748},
  2018.

\bibitem{chen2020simple}
T.~Chen, S.~Kornblith, M.~Norouzi, and G.~Hinton, ``A simple framework for
  contrastive learning of visual representations,'' \emph{arXiv preprint
  arXiv:2002.05709}, 2020.

\bibitem{srinivas2020curl}
A.~Srinivas, M.~Laskin, and P.~Abbeel, ``Curl: Contrastive unsupervised
  representations for reinforcement learning,'' \emph{arXiv preprint
  arXiv:2004.04136}, 2020.

\bibitem{grill2020bootstrap}
J.-B. Grill, F.~Strub, F.~Altch{\'e}, C.~Tallec, P.~H. Richemond,
  E.~Buchatskaya, C.~Doersch, B.~A. Pires, \emph{et~al.}, ``Bootstrap your own
  latent: A new approach to self-supervised learning,'' \emph{arXiv preprint
  arXiv:2006.07733}, 2020.

\bibitem{higgins2017darla}
I.~Higgins, A.~Pal, A.~Rusu, L.~Matthey, C.~Burgess, A.~Pritzel, M.~Botvinick,
  C.~Blundell, and A.~Lerchner, ``{DARLA}: Improving zero-shot transfer in
  reinforcement learning,'' \emph{ICML}, 2017.

\bibitem{tassa2018deepmind}
Y.~Tassa, Y.~Doron, A.~Muldal, T.~Erez, Y.~Li, D.~d.~L. Casas, D.~Budden,
  A.~Abdolmaleki, J.~Merel, A.~Lefrancq, \emph{et~al.}, ``Deepmind control
  suite,'' \emph{arXiv preprint arXiv:1801.00690}, 2018.

\bibitem{kaiser2019model}
L.~Kaiser, M.~Babaeizadeh, P.~Milos, B.~Osinski, R.~H. Campbell, K.~Czechowski,
  D.~Erhan, C.~Finn, P.~Kozakowski, S.~Levine, \emph{et~al.}, ``Model-based
  reinforcement learning for atari,'' \emph{arXiv:1903.00374}, 2019.

\bibitem{riedmiller2018learning}
M.~Riedmiller, R.~Hafner, T.~Lampe, M.~Neunert, J.~Degrave, T.~Van~de Wiele,
  V.~Mnih, \emph{et~al.}, ``Learning by playing-solving sparse reward tasks
  from scratch,'' \emph{arXiv preprint arXiv:1802.10567}, 2018.

\bibitem{wulfmeier2019regularized}
M.~Wulfmeier, A.~Abdolmaleki, R.~Hafner, J.~T. Springenberg, M.~Neunert,
  T.~Hertweck, T.~Lampe, N.~Siegel, N.~Heess, and M.~Riedmiller, ``Regularized
  hierarchical policies for compositional transfer in robotics,'' \emph{arXiv
  preprint arXiv:1906.11228}, 2019.

\bibitem{teh2017distral}
Y.~W. Teh, V.~Bapst, W.~M. Czarnecki, J.~Quan, J.~Kirkpatrick, R.~Hadsell,
  N.~Heess, and R.~Pascanu, ``Distral: Robust multitask reinforcement
  learning,'' \emph{CoRR}, vol. abs/1707.04175, 2017.

\bibitem{sharma2019dynamics}
A.~Sharma, S.~Gu, S.~Levine, V.~Kumar, and K.~Hausman, ``Dynamics-aware
  unsupervised discovery of skills,'' \emph{arXiv preprint arXiv:1907.01657},
  2019.

\bibitem{hertweck2020simple}
T.~Hertweck, M.~Riedmiller, M.~Bloesch, J.~T. Springenberg, N.~Siegel,
  M.~Wulfmeier, R.~Hafner, and N.~Heess, ``Simple sensor intentions for
  exploration,'' \emph{arXiv preprint arXiv:2005.07541}, 2020.

\bibitem{grimm2019disentangled}
C.~Grimm, I.~Higgins, A.~Barreto, D.~Teplyashin, M.~Wulfmeier, T.~Hertweck,
  R.~Hadsell, \emph{et~al.}, ``Disentangled cumulants help successor
  representations transfer to new tasks,'' \emph{arXiv preprint
  arXiv:1911.10866}, 2019.

\bibitem{relevantfreatures20}
J.~Luketina, M.~Smith, M.~Igl, and S.~Whiteson, ``Transfer learning via diverse
  policies in value-relevant features,'' \emph{BeTR-RL workshop, ICLR}, 2020.

\bibitem{laversannefinot2018curiosity}
A.~Laversanne-Finot, A.~Péré, and P.-Y. Oudeyer, ``Curiosity driven
  exploration of learned disentangled goal spaces,'' \emph{arXiv preprint
  arXiv:1807.01521}, 2018.

\bibitem{nair2018visual}
A.~V. Nair, V.~Pong, M.~Dalal, S.~Bahl, S.~Lin, and S.~Levine, ``Visual
  reinforcement learning with imagined goals,'' in \emph{NeurIPS}, 2018, pp.
  9191--9200.

\bibitem{higgins2018towards}
I.~Higgins, D.~Amos, D.~Pfau, S.~Racaniere, L.~Matthey, D.~Rezende, and
  A.~Lerchner, ``Towards a definition of disentangled representations,''
  \emph{arXiv preprint arXiv:1812.02230}, 2018.

\bibitem{bengio2017independently}
E.~Bengio, V.~Thomas, J.~Pineau, D.~Precup, and Y.~Bengio, ``Independently
  controllable features,'' \emph{arXiv preprint arXiv:1703.07718}, 2017.

\bibitem{abdolmaleki2018relative}
A.~Abdolmaleki, J.~T. Springenberg, J.~Degrave, S.~Bohez, Y.~Tassa, D.~Belov,
  N.~Heess, and M.~Riedmiller, ``Relative entropy regularized policy
  iteration,'' \emph{arXiv preprint arXiv:1812.02256}, 2018.

\bibitem{rezende2014stochastic}
D.~J. Rezende, S.~Mohamed, and D.~Wierstra, ``Stochastic backpropagation and
  approximate inference in deep generative models,'' \emph{arXiv preprint
  arXiv:1401.4082}, 2014.

\bibitem{bingham2001random}
E.~Bingham and H.~Mannila, ``Random projection in dimensionality reduction:
  applications to image and text data,'' in \emph{Proceedings of the seventh
  ACM SIGKDD}, 2001, pp. 245--250.

\bibitem{sutton1998introduction}
R.~S. Sutton, A.~G. Barto, \emph{et~al.}, \emph{Introduction to reinforcement
  learning}.\hskip 1em plus 0.5em minus 0.4em\relax MIT press Cambridge, 1998,
  vol. 135.

\bibitem{andrychowicz2017hindsight}
M.~Andrychowicz, D.~Crow, A.~Ray, J.~Schneider, R.~Fong, P.~Welinder,
  B.~McGrew, J.~Tobin, P.~Abbeel, and W.~Zaremba,
  ``\href{http://papers.nips.cc/paper/7090-hindsight-experience-replay.pdf}{Hindsight
  experience replay},'' in \emph{NeurIPS}, 2017, pp. 5055--5065.

\bibitem{nachum2018data}
O.~Nachum, S.~S. Gu, H.~Lee, and S.~Levine, ``Data-efficient hierarchical
  reinforcement learning,'' in \emph{NeurIPS}, 2018, pp. 3303--3313.

\bibitem{yu2020meta}
T.~Yu, D.~Quillen, Z.~He, R.~Julian, K.~Hausman, C.~Finn, and S.~Levine,
  ``Meta-world: A benchmark and evaluation for multi-task and meta
  reinforcement learning,'' in \emph{CoRL}, 2019.

\bibitem{donahue2019large}
J.~Donahue and K.~Simonyan, ``Large scale adversarial representation
  learning,'' in \emph{NeurIPS}, 2019.

\bibitem{mnih2013playing}
V.~Mnih, K.~Kavukcuoglu, D.~Silver, A.~Graves, I.~Antonoglou, D.~Wierstra, and
  M.~Riedmiller, ``Playing atari with deep reinforcement learning,''
  \emph{arXiv preprint arXiv:1312.5602}, 2013.

\bibitem{lillicrap2015continuous}
T.~P. Lillicrap, J.~J. Hunt, A.~Pritzel, N.~Heess, T.~Erez, Y.~Tassa,
  D.~Silver, and D.~Wierstra, ``Continuous control with deep reinforcement
  learning,'' in \emph{ICLR}, 2016.

\bibitem{kalashnikov2018qt}
D.~Kalashnikov, A.~Irpan, P.~Pastor, J.~Ibarz, A.~Herzog, E.~Jang, D.~Quillen,
  \emph{et~al.}, ``Qt-opt: Scalable deep reinforcement learning for
  vision-based robotic manipulation,'' \emph{arXiv preprint arXiv:1806.10293},
  2018.

\bibitem{byravan2020imagined}
A.~Byravan, J.~T. Springenberg, A.~Abdolmaleki, R.~Hafner, M.~Neunert,
  \emph{et~al.}, ``Imagined value gradients: Model-based policy optimization
  with transferable latent dynamics models,'' in \emph{CoRL}, 2019, pp.
  566--589.

\bibitem{hafner2019dream}
D.~Hafner, T.~Lillicrap, J.~Ba, and M.~Norouzi, ``Dream to control: Learning
  behaviors by latent imagination,'' \emph{arXiv preprint arXiv:1912.01603},
  2019.

\bibitem{lee2019stochastic}
A.~X. Lee, A.~Nagabandi, P.~Abbeel, and S.~Levine, ``Stochastic latent
  actor-critic: Deep reinforcement learning with a latent variable model,''
  \emph{arXiv preprint arXiv:1907.00953}, 2019.

\bibitem{locatello2019challenging}
F.~Locatello, S.~Bauer, M.~Lucic, G.~Raetsch, S.~Gelly, B.~Sch{\"o}lkopf, and
  O.~Bachem, ``Challenging common assumptions in the unsupervised learning of
  disentangled representations,'' in \emph{ICML}, 2019, pp. 4114--4124.

\bibitem{schwarzer2020data}
M.~Schwarzer, A.~Anand, R.~Goel, R.~D. Hjelm, A.~Courville, and P.~Bachman,
  ``Data-efficient reinforcement learning with momentum predictive
  representations,'' \emph{arXiv preprint arXiv:2007.05929}, 2020.

\bibitem{lange2012autonomous}
S.~Lange, M.~Riedmiller, and A.~Voigtl{\"a}nder, ``Autonomous reinforcement
  learning on raw visual input data in a real world application,'' in
  \emph{IJCNN}.\hskip 1em plus 0.5em minus 0.4em\relax IEEE, 2012, pp. 1--8.

\bibitem{aytar2018playing}
Y.~Aytar, T.~Pfaff, D.~Budden, T.~Paine, Z.~Wang, and N.~de~Freitas, ``Playing
  hard exploration games by watching youtube,'' in \emph{NeurIPS}, 2018, pp.
  2930--2941.

\bibitem{Dosovitskiy2017}
A.~Dosovitskiy and V.~Koltun, ``Learning to act by predicting the future,'' in
  \emph{ICLR}, 2017.

\bibitem{Jaderberg2017Unreal}
M.~Jaderberg, V.~Mnih, W.~M. Czarnecki, T.~Schaul, J.~Z. Leibo, D.~Silver, and
  K.~Kavukcuoglu, ``\href{https://arxiv.org/pdf/1611.05397.pdf}{Unreal}:
  Reinforcement learning with unsupervised auxiliary tasks,'' in \emph{ICLR},
  2017.

\bibitem{Mirowski16}
P.~Mirowski, R.~Pascanu, F.~Viola, H.~Soyer, A.~J. Ballard, A.~Banino,
  M.~Denil, R.~Goroshin, L.~Sifre, \emph{et~al.},
  ``\href{http://arxiv.org/abs/1611.03673}{Learning to Navigate} in complex
  environments,'' \emph{arXiv preprint arXiv:1611.03673}, 2016.

\bibitem{cabi2017}
S.~Cabi, S.~G. Colmenarejo, M.~W. Hoffman, M.~Denil, Z.~Wang, and
  N.~de~Freitas,
  ``\href{http://proceedings.mlr.press/v78/cabi17a/cabi17a.pdf}{The Intentional
  Unintentional Agent}: Learning to solve many continuous control tasks
  simultaneously,'' in \emph{CoRL}, 2017.

\bibitem{kaelbling1993learning}
L.~P. Kaelbling, ``Learning to achieve goals,'' in \emph{IJCAI}.\hskip 1em plus
  0.5em minus 0.4em\relax Citeseer, 1993, pp. 1094--1099.

\bibitem{wulfmeier2020data}
M.~Wulfmeier, D.~Rao, R.~Hafner, T.~Lampe, A.~Abdolmaleki, T.~Hertweck,
  M.~Neunert, D.~Tirumala, N.~Siegel, N.~Heess, \emph{et~al.}, ``Data-efficient
  hindsight off-policy option learning,'' \emph{arXiv preprint
  arXiv:2007.15588}, 2020.

\bibitem{eysenbach2018diversity}
B.~Eysenbach, A.~Gupta, J.~Ibarz, and S.~Levine, ``Diversity is all you need:
  Learning skills without a reward function,'' \emph{arXiv preprint
  arXiv:1802.06070}, 2018.

\bibitem{pathak2017curiosity}
D.~Pathak, P.~Agrawal, A.~A. Efros, and T.~Darrell, ``Curiosity-driven
  exploration by self-supervised prediction,'' in \emph{IEEE CVPR Workshops},
  2017, pp. 16--17.

\bibitem{Oudeyer17}
S.~Forestier, Y.~Mollard, and P.~Oudeyer, ``Intrinsically motivated goal
  exploration processes with automatic curriculum learning,'' \emph{arXiv
  preprint arXiv:1708.02190}, 2017.

\bibitem{klyubin2005all}
A.~S. Klyubin, D.~Polani, and C.~L. Nehaniv, ``All else being equal be
  empowered,'' in \emph{European Conference on Artificial Life}.\hskip 1em plus
  0.5em minus 0.4em\relax Springer, 2005, pp. 744--753.

\bibitem{gregor2016variational}
K.~Gregor, D.~J. Rezende, and D.~Wierstra, ``Variational intrinsic control,''
  in \emph{ICLR}, 2017.

\bibitem{graves2017automated}
A.~Graves, M.~G. Bellemare, J.~Menick, R.~Munos, and K.~Kavukcuoglu,
  ``Automated curriculum learning for neural networks,'' in \emph{ICML}.\hskip
  1em plus 0.5em minus 0.4em\relax JMLR. org, 2017, pp. 1311--1320.

\bibitem{bengio2009curriculum}
Y.~Bengio, J.~Louradour, R.~Collobert, and J.~Weston, ``Curriculum learning,''
  in \emph{ICML}, 2009.

\bibitem{heess2017emergence}
N.~Heess, D.~Tirumala, S.~Sriram, J.~Lemmon, J.~Merel, G.~Wayne, Y.~Tassa,
  T.~Erez, Z.~Wang, \emph{et~al.}, ``Emergence of locomotion behaviours in rich
  environments,'' \emph{arXiv preprint arXiv:1707.02286}, 2017.

\bibitem{florensa2017reverse}
C.~Florensa, D.~Held, M.~Wulfmeier, M.~Zhang, and P.~Abbeel, ``Reverse
  curriculum generation for reinforcement learning,'' \emph{arXiv preprint
  arXiv:1707.05300}, 2017.

\bibitem{Schmidhuber2013PowerPlayTA}
J.~Schmidhuber, ``\href{https://arxiv.org/pdf/1112.5309.pdf}{PowerPlay}:
  Training an increasingly general problem solver by continually searching for
  the simplest still unsolvable problem,'' in \emph{Front. Psychol.}, 2013.

\bibitem{wang2019paired}
R.~Wang, J.~Lehman, \emph{et~al.}, ``Paired open-ended trailblazer (poet):
  Endlessly generating increasingly complex and diverse learning environments
  and their solutions,'' \emph{arXiv preprint arXiv:1901.01753}, 2019.

\bibitem{todorov2012mujoco}
E.~Todorov, T.~Erez, and Y.~Tassa, ``Mujoco: A physics engine for model-based
  control,'' in \emph{Intelligent Robots and Systems (IROS), 2012 IEEE/RSJ
  International Conference on}.\hskip 1em plus 0.5em minus 0.4em\relax IEEE,
  2012, pp. 5026--5033.

\bibitem{jeong2019self}
R.~Jeong, Y.~Aytar, D.~Khosid, Y.~Zhou, J.~Kay, T.~Lampe, K.~Bousmalis, and
  F.~Nori, ``Self-supervised sim-to-real adaptation for visual robotic
  manipulation,'' \emph{arXiv preprint arXiv:1910.09470}, 2019.

\bibitem{watters2019}
N.~Watters, L.~Matthey, C.~P. Burgess, and A.~Lerchner, ``Spatial broadcast
  decoder: A simple architecture forlearning disentangled representations in
  vaes,'' \emph{arXiv preprint arXiv:1901.07017}, 2019.

\end{thebibliography}

\clearpage
\appendix
\section*{Generalization to Physical Robot Experiments}

This submission purely focuses on experiments in simulations as due to the ongoing pandemic robot access is highly limited for us.
We will use the following sections to address any concerns regarding generalization to the real-world case by given examples of our previous work and general experimental setup. 
Any experiments and results in this section are not part of the contributions of this paper and serve the only purpose to clarify generality and robustness of results in the main paper.
The following sections highlight 3 principal steps we take in order to support generality of the main paper's simulation results.

\subsection*{Realism of Simulator}

The experiments described in the main section of this paper were conducted with a carefully designed MuJoCo \cite{todorov2012mujoco} simulation that is visually and dynamically well aligned with the real world robot setup. This simulation was successfully used in Sim2Real transfer before \cite{jeong2019self}, giving strong evidence that the methods presented in this paper can be applied successfully to the real world robot setup as well.

\subsection*{Representations from Real-World Data}

\begin{figure}[h]
\begin{minipage}[b]{.16\linewidth}
\centering
\begin{tabular}{c}
    \includegraphics[width=.7875\textwidth]{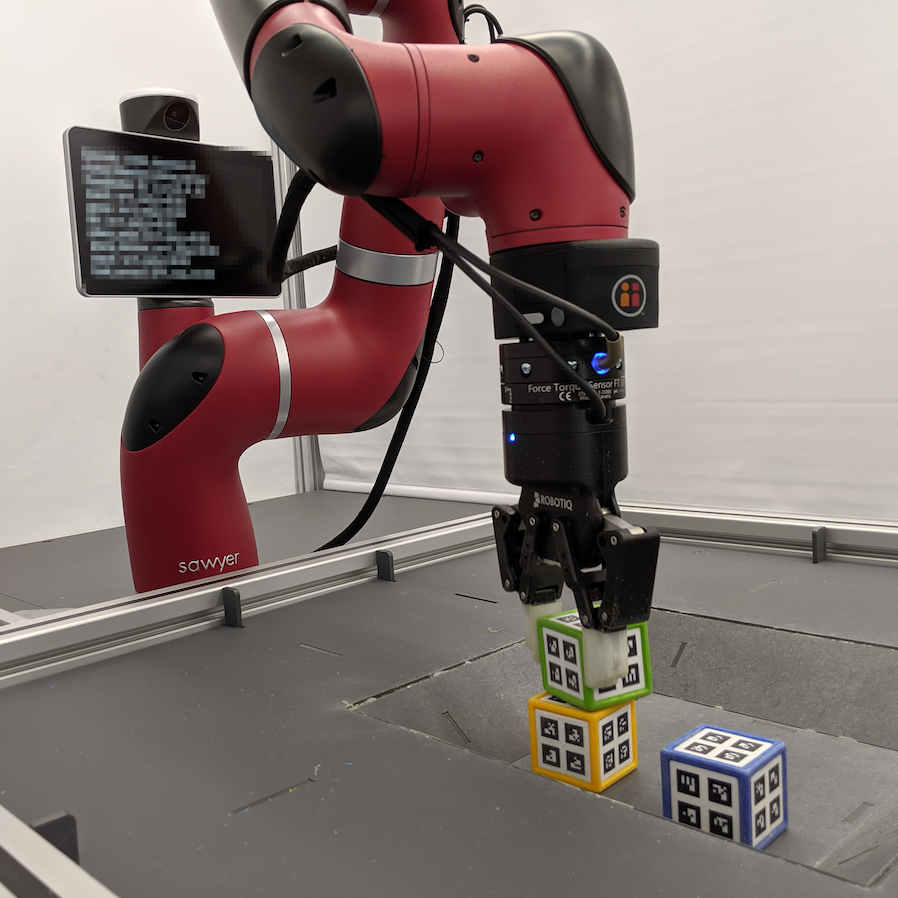}\\
    \includegraphics[width=.7875\textwidth]{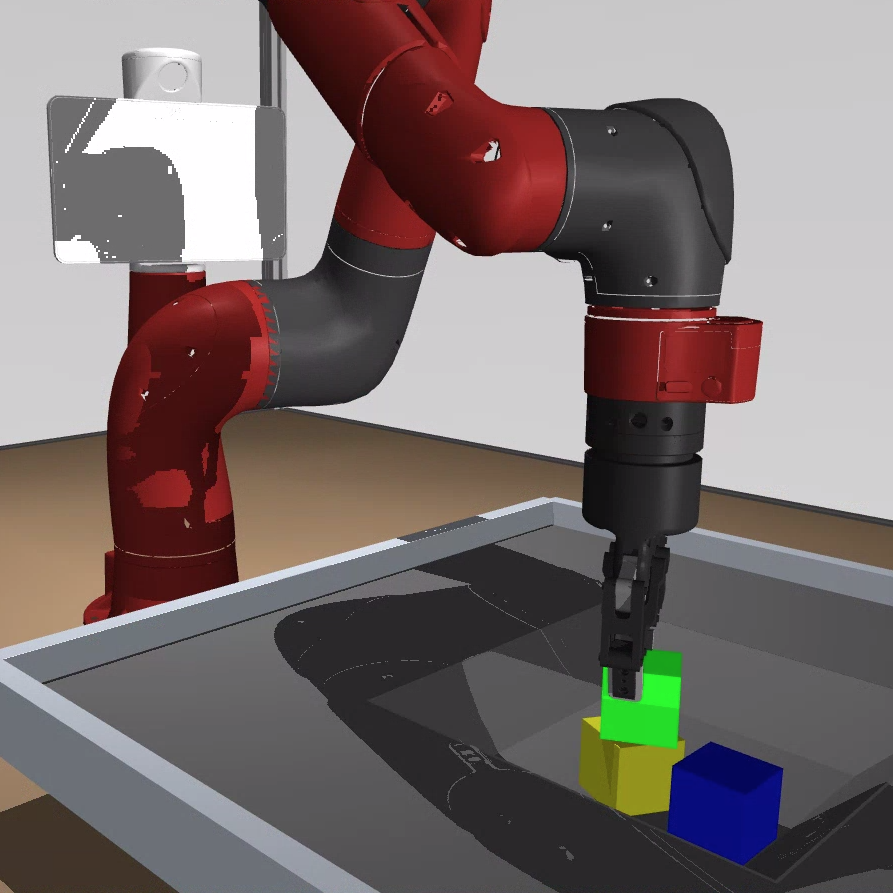}
\end{tabular}
\end{minipage}%
\begin{minipage}[b]{.28\linewidth}
\centering
\begin{tabular}{cc}
    \includegraphics[width = 0.45\textwidth]{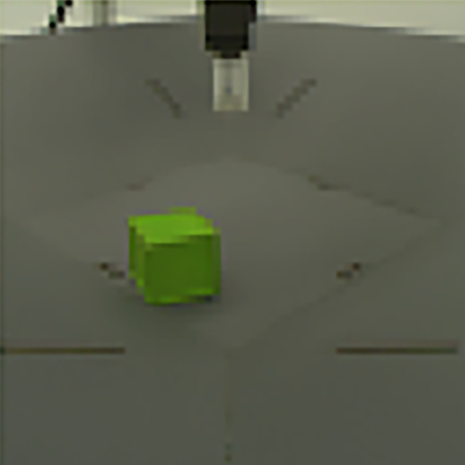}
    \includegraphics[width = 0.45\textwidth]{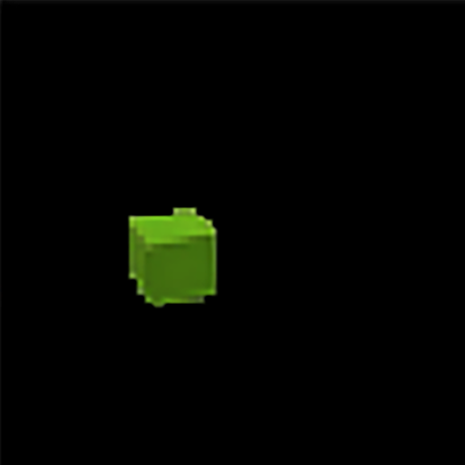} \\
    \includegraphics[width = 0.45\textwidth]{images/blob_sim_input.png}
    \includegraphics[width = 0.45\textwidth]{images/blob_sim_mask.png}
\end{tabular}
\end{minipage}%
\begin{minipage}[b]{.28\linewidth}
\centering
\begin{tabular}{c}
    \includegraphics[trim=4.5cm 4.5cm 0 2.25cm,clip,width = 0.9\textwidth]{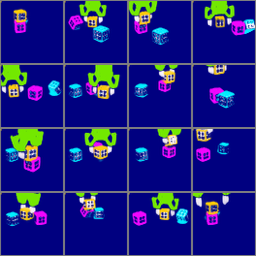} \\
    \includegraphics[trim=4.5cm 4.5cm 0 2.25cm,clip,width = 0.9\textwidth]{images/image6-small-19.png}
\end{tabular}
\end{minipage}%
\begin{minipage}[b]{.28\linewidth}
\centering
\begin{tabular}{cc}
    \includegraphics[width = 0.45\textwidth]{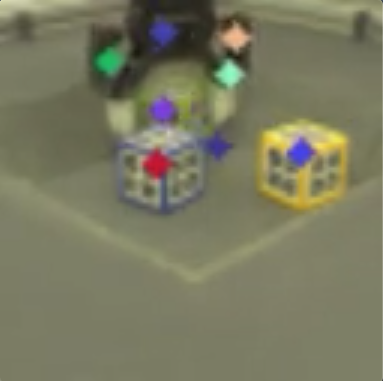} 
    \includegraphics[width = 0.45\textwidth]{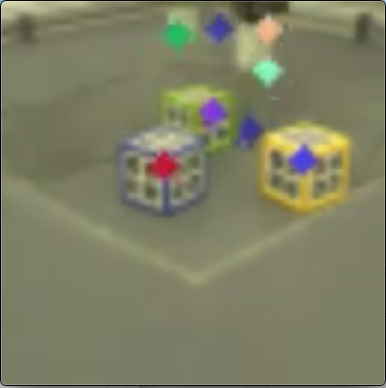} \\
    \includegraphics[width = 0.45\textwidth,height = 0.45\textwidth]{images/t_sim_skew1.png}
    \includegraphics[width = 0.45\textwidth,height = 0.45\textwidth]{images/t_sim_skew2.png} 
\end{tabular}
\end{minipage}

\caption{Representations obtained from simulated and real world datasets. Left: The robot setups. Mid-Left: Blob Detector outputs. Mid-Right: MONet segmentations, Right: Transporter key-points. Top: Real world data. Bottom: Simulated data.}
\label{fig:simandreal}
\end{figure}

Even though we did not have the possibility to conduct experiments on a real world robot setup, we are able to compare the representation learning methods on simulated and real world data, which was collected before the pandemic. As shown in Figure \ref{fig:simandreal}, both representation learning methods employed in the main section of this paper transfer well to real world data, giving evidence that using the learned representation for state and reward computation is beneficial for real world experiments.

\subsection*{Connection to Related Work}

\begin{figure}[h]
    \centering
    \includegraphics[width=0.5\textwidth]{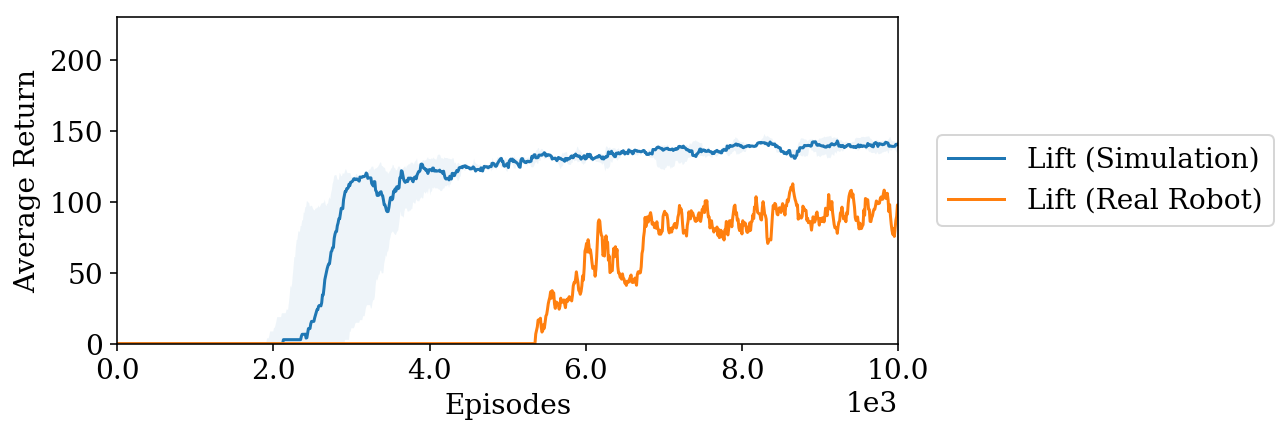}
    \caption{`Lift' learned in simulation and on a real robot \cite{hertweck2020simple}.}
    \label{fig:lift_sim_real}
\end{figure}

Further, we want to refer to previous work that was conducted with Scheduled-Auxiliary-Control (SAC-X) on this robot setup, both in simulation and in the real world. In \cite{hertweck2020simple}, the authors use the color segmentation approach outlined in the main section of this paper for reward computation. Figure \ref{fig:lift_sim_real} shows, that SAC-X is comparably capable of learning the lifting task in both setups. In \cite{wulfmeier2019regularized}, the authors show results for the stacking task both in simulation and on the real robot, giving further evidence that learning based on SAC-X transfers well to real-world robot setups.

\section*{Experiment Details} \label{app:taskdetails}

\paragraph{Sawyer Environment}
The setup consists of a Sawyer robot arm mounted on a table and equipped with a Robotiq 2F-85 parallel gripper. In front of the robot there is a basket of size 20x20 cm which contains three cubes with an edge length of 5 cm (see Figure \ref{fig:robot}).

The agent is provided with proprioception information for the arm (joint positions, velocities and torques), and the tool center point position computed via forward kinematics. For the gripper, it receives the motor position and velocity, as well as a binary grasp flag. It also receives a wrist sensor's force and torque readings. Finally, it is provided with RGB camera images at $64 \times 64$ resolution. At each time step, a history of two previous observations (except for the images) is provided to the agent, along with the last two joint control commands. The observation space is detailed in Table \ref{tab:sawyer_observations}. All experiments are run with 16 actors in parallel and reported over the current episodes generated by any actor.

The robot arm is controlled in Cartesian velocity mode at 20Hz. The action space for the agent is 5-dimensional, as detailed in Table \ref{tab:pile_actions}. The gripper movement is also restricted to a cubic volume above the basket using virtual walls.
\begin{table}[h!]
\caption{Action space for the Sawyer Stacking experiments.}
\label{sawyer-action-table2}
\vskip 0.15in
\begin{center}
\begin{small}
\begin{tabular}{lccc}
\toprule
Entry & Dims & Unit & Range \\
\midrule
Translational Velocity in x, y, z & 3 & m/s & [-0.07, 0.07] \\
Wrist Rotation Velocity & 1 & rad/s & [-1, 1] \\
Finger speed & 1 & tics/s & [-255, 255] \\
\bottomrule
\end{tabular}
\end{small}
\end{center}
\vskip -0.1in
\label{tab:pile_actions}
\end{table}

\begin{table}[h!]
\caption{Observations for the experiments. The TCP's pose is represented as its world coordinate position and quaternion. In the table, $m$ denotes meters, $rad$ denotes radians, and $q$ refers to a quaternion in arbitrary units ($au$).}
\label{tab:sawyer_observations}
\vskip 0.15in
\begin{center}
\begin{small}
\begin{tabular}{lccc}
\toprule
Entry & Dims & Unit & History \\
\midrule
Joint Position (Arm) & 7 & rad & 2 \\
Joint Velocity (Arm) & 7 & rad/s & 2 \\
Joint Torque (Arm) & 7 & Nm & 2 \\
Joint Position (Hand) & 1 & tics & 2 \\
Joint Velocity (Hand) & 1 & tics/s & 2 \\
Force-Torque (Wrist) & 6 & N, Nm & 2 \\
Binary Grasp Sensor & 1 & au & 2 \\
TCP Pose & 7 & m, au & 2 \\
Camera images & $3 \times 64 \times 64 \times 3$ & R/G/B value & 1 \\
Last Control Command & 8 & rad/s, tics/s & 2 \\
\bottomrule
\end{tabular}
\end{small}
\end{center}
\vskip -0.1in
\end{table}

\paragraph{Tasks and Rewards}
We use shaped rewards for specifying auxiliary tasks for the experiments where we use the learned representations as inputs to the agent. These are not used for the task generation experiments where auxiliary rewards are based on the learned representations themselves. Below, we discuss the reward setup for the three major tasks considered in the paper, the \textbf{Lift}, \textbf{Stack} and \textbf{Match Positions} task. 

\textbf{Lift: } In the lift task, the agent has to pick-up the green block on the table and lift it above a certain height. We introduce additional auxiliary tasks that encourage reaching the target object, grasping it and the final task of lifting. These are specified in turn as:

\begin{itemize}
    \item \textit{REACH(G)}: $stol(d(TCP, G), 0.02, 0.15)$: \\
    Minimize the distance of the TCP to the green cube.
    \item \textit{GRASP}: \\
    Activate grasp sensor of gripper ("inward grasp signal" of Robotiq gripper)
    \item \textit{LIFT(G)}: $slin(G, 0.03, 0.10)$ \\
    Increase z coordinate of the green cube by more than 3cm relative to the table.
\end{itemize}

Where $d(x,y)$ is the Euclidean distance between a pair of 3D points, and the tolerance and linear reward functions terms, respectively $stol$ and $slin$, are defined as:
\begin{equation}
stol(v, \epsilon, r) =
\begin{cases}
  1 &\text{iff} \ |v| < \epsilon \\
  1 - tanh^2( \frac{atanh(\sqrt{0.95})}{r} |v|) &\text{else}
\end{cases}
\label{eq:shaped_tolerance}
\end{equation}

\begin{equation}
slin(v, \epsilon_{min}, \epsilon_{max}) =
\begin{cases}
  0 &\text{iff} \ v < \epsilon_{min} \\
  1 &\text{iff} \ v > \epsilon_{max} \\
  \frac{v - \epsilon_{min}}{\epsilon_{max} - \epsilon_{min}}  &\text{else}
\end{cases}
\label{eq:shaped_tolerance2}
\end{equation}

\textbf{Stack:} In this task, the agent's final task is to stack the green cube on top of the yellow cube. We again introduce auxiliary tasks that encouraging the agent to lift the object -- all auxiliary tasks and the final lift task are used as auxiliary tasks for stacking. The additional auxiliary tasks are specified to encourage the agent to move towards the target, align it with the target block and stack the grasped object:

\begin{itemize}
    \item \textit{PLACE\_WIDE(G, Y) - stack domain}: $stol(d(G, Y + [0,0,0.05]), 0.01, 0.20)$\\
    Bring green cube to a position 5cm above the yellow cube.
    \item \textit{PLACE\_NARROW(G, Y) - stack domain}: $stol(d(G, Y + [0,0,0.05]), 0.00, 0.01)$: \\
    Like PLACE\_WIDE(G, Y) but more precise.
    \item \textit{STACK(G, Y) - stack domain}: $btol(d_{xy}(G, Y), 0.03) * btol(d_z(G, Y) + 0.05, 0.01) * (1 - \textit{GRASP})$ \\
    Sparse binary reward for bringing the green cube on top of the yellow one (with 3cm tolerance horizontally and 1cm vertically) and disengaging the grasp sensor.
\end{itemize}

Where the tolerance function $btol$ is defined as:
\begin{equation}
btol(v, \epsilon) =
\begin{cases}
  1 &\text{iff} \ |v| < \epsilon  \\
  0 &\text{else}
\end{cases}
\end{equation}

\textbf{Match Positions: } In this task, the agent has to move both the green and yellow cubes to a fixed target position. 
The task involves moving both objects and enables in part testing the generalization of our learned representations which were trained on data where the arm only interacts with the primary object of interest i.e. the green cube. Additionally, the final reward for each block is sparse but reward is individually received for each block crossing a distance threshold to the respective targets ($t1,t2$). Auxiliary rewards are given for reach tasks for each block.
We specify the rewards as:
\begin{itemize}
    \item \textit{REACH(G)}: $stol(d(TCP, G), 0.02, 0.15)$: \\
    Minimize the distance of the TCP to the green cube.
    \item \textit{REACH(Y)}: $stol(d(TCP, Y), 0.02, 0.15)$: \\
    Minimize the distance of the TCP to the yellow cube.
    \item \textit{MATCH\_POSITIONS(G,Y)}: $btol(d(G,t1),0.02) + btol(d(Y,t2),0.02)$
\end{itemize}

where $t1, t2$ denote the target 3D positions of the green and yellow block respectively.

\paragraph{Dataset}
The dataset to train all representation learning models was generated from the an agent which was trained to solved all 6 tasks described above, we train the agent with simulator states as inputs to accelerate training but log rendered images. The dataset includes about $3*10^7$ transitions in 52150 episodes of about length 600 steps.

\section*{Method Details} \label{app:methoddetails}

\subsection*{VAE and \ensuremath{\beta}-VAE}
In practice, our VAE implementation consists of an inference network, that takes in images $I$ and parameterizes the posterior distribution $q(Z|I)$, and a generative network that takes a sample from the inferred posterior distribution and attempts to reconstruct the original image. The model is trained through a two-part loss objective:
\begin{equation}
\begin{aligned}
    \mathcal{L}_{VAE} = \mathbb{E}_{p(\mathbf{x})} [\ \mathbb{E}_{q_{\phi}(\mathbf{z}|\mathbf{x})} [\log\ p_{\theta}(\mathbf{x} | \mathbf{z})] - \\ 
    \nonumber KL(q_{\phi}(\mathbf{z}|\mathbf{x})\ ||\ p(\mathbf{z}))\ ]
\end{aligned}
\end{equation}
where $p(\mathbf{x})$ is the probability of the image data, $q(\mathbf{z}|\mathbf{x})$ is the learnt posterior over the latent units given the data, and $p(\mathbf{z})$ is the unit Gaussian prior with a diagonal covariance matrix. 
Intuitively, the objective consists of the reconstruction term (which aims to increase the log-likelihood of the image observations) and a compression term (which aims to reduce the divergence between the prior, normally chosen to be an isotropic unit Gaussian, and the inferred posterior). We use the means from the posterior inferred by the pre-trained encoder network to produce the embedding $e$.
\paragraph{$\beta$-VAE}\label{sec:bvae}

A $\beta$-VAE \cite{higgins2017beta} is a variation of the VAE model, where the second part of the training objective is augmented with a $\beta$ hyperparameter.
\begin{equation}
\begin{aligned}
    \mathcal{L}_{\ensuremath{\beta}-VAE} = \mathbb{E}_{p(\mathbf{x})} [\ \mathbb{E}_{q_{\phi}(\mathbf{z}|\mathbf{x})} [\log\ p_{\theta}(\mathbf{x} | \mathbf{z})] - 
    \\ \nonumber \beta KL(q_{\phi}(\mathbf{z}|\mathbf{x})\ ||\ p(\mathbf{z}))\ ]
\end{aligned}
\end{equation}
The $\beta$ hyperparameter controls the degree of disentangling achieved by the model during training. Typically a $\beta>1$ is necessary to achieve good disentangling, however the exact value differs for different datasets. To obtain entangled representations a low $\beta$ is chosen for our experiments.

We used the standard architecture and optimization parameters introduced in \cite{higgins2017beta} for training the \ensuremath{\beta}-VAE models. The encoder consisted of four convolutional layers (all 32x4x4 stride 2), followed by a 128-d fully connected layer and a 2-d, 16-d or a 32-d latent representation, depending on the desired dimensionality of the embedding. The decoder architecture was the reverse of the encoder. We used ReLU activations throughout. The decoder parametrized a Bernoulli distribution. We used Adam optimizer with $1e-4$ learning rate and trained the models for  about $10^6$ iterations using batch size of 16, which was enough to achieve convergence. All final models used were able to reconstruct all objects in the images. Images were presented at 128x128x2 resolution.

\subsection*{MONet}
MONet \cite{burgess2019monet} is an unsupervised method for discovering object-based representations of visual scenes. It augments a VAE (or a $\beta$-VAE) with a recurrent attention network, which decomposes the input image $I$ into $N$ slots based on the corresponding attention masks. The $N$ attention masks partition the original set of input pixels into $N$ sets, each of which corresponds to a meaningful entity in the visual scene (e.g. an object, or the background). These attention attenuated images are then fed into a shared VAE network applied sequentially to them, and the weighted sum of the respective reconstructions is trained to correspond to the original input image.  The attention network and the VAE are optimized end-to-end in an unsupervised manner. The concatenation of the $N$ embeddings obtained from the sequential application of the VAE to the $N$ attention-attenuated input images constitute the final MONet embedding $e$.

We used the standard MONet architecture as described in \cite{burgess2019monet}, with 6 object slots, and two different internal \ensuremath{\beta}-VAEs  -- one applied to the first slot, and the other to the remaining 5 slots. Both \ensuremath{\beta}-VAE models had a 16-d latent representation and used a broadcast decoder \cite{watters2019}. The model was trained for 1~mln iterations, and the only difference in hyperparameters from the original MONet is that we used $\beta=5$ to encourage disentangling within slots, and we set $\sigma_{fg} = 0.12$ for the scale of the "foreground" components.

\subsection*{Transporter}
Transporter~\cite{kulkarni2019unsupervised} is an unsupervised method for learning object keypoints or image-space 2D coordinates of independently moving entities from videos. The model learns by reconstructing a \emph{target} frame $x_t$ from another randomly sampled source frame $x_s$, by moving/transporting the features at the discovered keypoint locations in the source frame to their corresponding locations in the target frame. The discovered keypoint-coordinates (normalized to the [-1, 1] range) are used as the embedding. Refer to the original publication~\cite{kulkarni2019unsupervised} for a detailed description.

We followed the implementation in~\cite{kulkarni2019unsupervised} and used identical architectures for the keypoint and image encoders: a four layer 2D convolutional network with (32, 32, 64, 128) / (1, 1, 2, 1) / (7, 3, 3, 3) as the number of filters / stride / filter-size respectively, acting on $128{\times}128$-pixel RGB images. Standard-deviation of the keypoint Gaussians was set to 0.1,
and the number of keypoints was set to $10$.
The refinement network was the same as in the original work. We used the Adam optimizer with the gradient-norm clipped to 1.0, and an initial learning rate of 1e-3 decayed by 0.95 every 100k steps. The source and target frames were sampled randomly from buffers containing 5000 images each extracted from different episodes.

\subsection*{Color Segmentation}
In the following, we describe the transformation we use to obtain representations through color segmentation from raw camera images by aggregating statistics of an image's spatial color distribution.

\begin{figure}
    \centering
    \includegraphics[width=.9\linewidth]{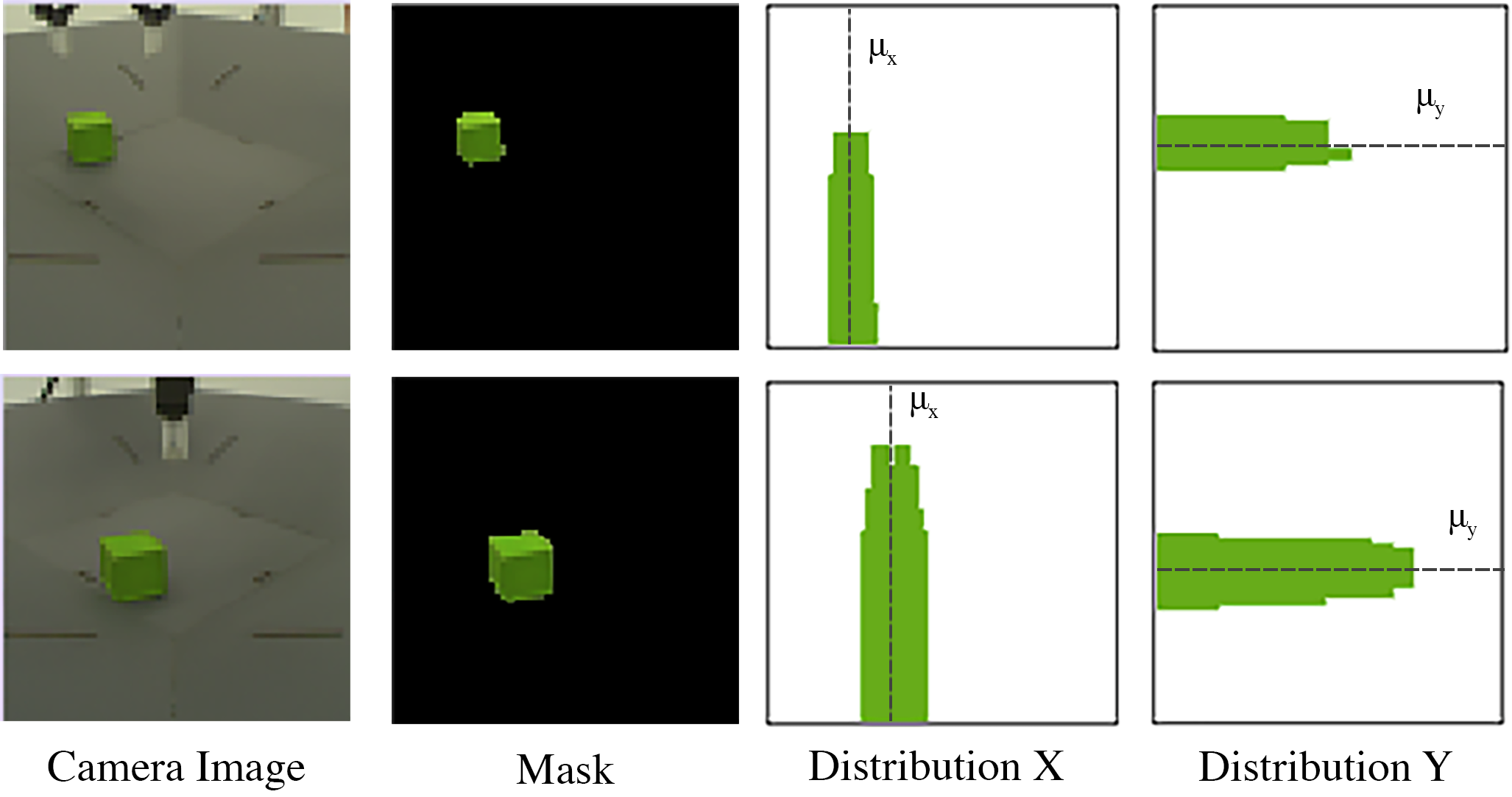}
    \caption{The transformation used for deriving representations from camera images. We compute a binary mask and it's spatial distribution along the image axes and select the resulting distribution's mean as a representation for the color range.}
    \label{fig:histogram}
\end{figure}

 As illustrated in Figure \ref{fig:histogram}, we first threshold the image to retain only a given color (resulting in a binary mask) and then calculate the mean location of the mask along each of the image's axes, which we use as a representation for the color range. Formally we can define the two corresponding representation dimensions for each camera image and color range as
\begin{equation*}
\begin{aligned}
z^{c_\text{range}}_x &= \frac{1}{W} \sum_{x=0}^W  x \max_y [\mathbf{1}_{c_\text{range}}(o_\text{image})[y, x]] \\
z^{c_\text{range}}_y &= \frac{1}{H} \sum_{x=0}^H  y \max_x [\mathbf{1}_{c_\text{range}}(o_\text{image})[y, x]],
\end{aligned}
\end{equation*}
where $H$ denotes the image height, $W$ the width, and $c_\text{range} = [c_\text{min}$, $c_\text{max}]$ correspond to color ranges that should be filtered.

The choice of the `right' color range $c_\text{range}$, for a task of interest, is a design decision that needs to be made manually. In practice, we define a set of color ranges (and corresponding sensor values) from rough estimates of the color of objects of interest in the scene.

\subsection*{Random Projection}
The simple random projection approach \cite{bingham2001random} we use builds on a randomly initialized matrix $M \in \mathbb{R}^{(h*w*3,d)}$ where $h$ and $w$ denote the image height and width respectively and $d$ is the embedding size. The embedding $z$ can be computed as $z = \text{tanh} (M \cdot \text{flatten}(I)) $, where we apply a saturating non-linearity (tanh) for bounding the output. 
The initialization follows a truncated Normal distribution with mean 0 and standard deviation $ 1 /\sqrt{\text{size of input layer}}$.

\subsection*{Task Generation}
For the task generation use-case, we estimate minimum and maximum of an embedding over the training dataset and in addition clip values that exceed the estimated extremes.
We define auxiliary tasks such that for each dimension $i$ of the embedding $z(x)$, we create two tasks, denoted $min$ and $max$, respectively for for minimizing and maximizing its value:
\begin{align}
    r_{i,max}(x,a) = 1 -\frac{\max_x(z_i(x)) - z_i(x)}{\text{range}(z_i)}\\
    r_{i,min}(x,a) = 1 - \frac{z_i(x) - \min_x(z_i(x))}{\text{range}(z_i)}\nonumber\\
    \text{with: } \text{range}(z_i) = \max_x(z_i(x)) - \min_x(z_i(x)) \nonumber
\end{align}
where $z_i$ denotes the $i$th dimension of $z$ and we normalize the rewards to make multi-task training robust. To simplify learning, the rewards are normalised by the maximum and minimum values of each embedding.

\newcommand{\bE}{\mathbb{E}}
\newcommand{\bx}{\mathbf{x}}
\newcommand{\bs}{\mathbf{s}}
\newcommand{\ba}{\mathbf{a}}
\newcommand{\btheta}{{\bm{\theta}}}
\newcommand{\cM}{{\mathcal{M}}}
\newcommand{\sA}{\mathscr{A}}
\newcommand{\sT}{\mathscr{T}}
\newcommand{\cA}{{\mathcal{A}}}
\newcommand{\cI}{{\mathcal{I}}}
\newcommand{\cT}{{\mathcal{T}}}
\newcommand{\cB}{{\mathcal{B}}}
\newcommand{\cL}{{\mathcal{L}}}
\newcommand{\cS}{{\mathcal{S}}}
\newcommand{\pluseq}{\mathrel{+}=}

\subsection*{Improved Q-Scheduler}
As mentioned in the main paper, we use a task scheduling procedure that is based on the Q-scheduler in \cite{riedmiller2018learning} (SAC-Q), which we will slightly improve.

In \cite{riedmiller2018learning} the scheduling scheme is described to have two almost independent phases.
In the first phase, when no reward for the main task has been observed, a purely random schedule is used to interact with the environment. After some reward for the main task was discovered, scheduling changes to increase sequence combinations of tasks proportional to the main task reward which can be observed by executing them.

We use the very same basic principle, but improve the initial phase by replacing random scheduling with a more data-efficient procedure.
The main idea of this improvement is that during the first phase in each episode, we select one of the auxiliary tasks to be used as the main task reward for the SAC-Q scheduling approach instead.
By using an extended table representation for all of the auxiliaries, instead of only the main task, we can now decide which combination of intentions to execute in order to make learning of the selected auxiliary as efficient as possible. Intuitively, the approach aims to accelerate learning progress on any task as long as the main task provides no rewards. 
For the decision which of the auxiliary tasks to select as the replacement main task, we use multiple heuristics.
First, we don't select auxiliary tasks for that we never saw any reward signal, analogous to the arguments why we wouldn't use the main task.
Second, we don't want to waste interaction data by selecting auxiliaries for which we already see high reward at the initial state of the episode (threshold determined by the rewards seen for this auxiliary during previous executions).
This already prevents to e.g. select a "reach" task when the gripper is already close to the object.
From the remaining candidate set, we sample with a probability that is proportional to the increase of average reward that was seen (while executing any intention policy) for that auxiliary.

After we selected one of the auxiliaries, $\cI$, to be treated as the main task at the beginning of an episode, we can use the scheme of SAC-Q to determine the probabilities $P_{\cS}$ for the tasks that we want to execute in the sequences of this episode following equation \ref{eq:softmax_sched}:
\begin{equation}
P_\cS(\cT_h | \cT_{1:h-1}; \eta) = \frac{\exp(\bE_{P_\cS}[R_\cI(\cT_{h:H})] / \eta)}{\sum_{\bar{\cT}_{h:H}} \exp(\bE_{P_\cS}\lbrack R_\cI(\bar{\cT}_{h:H})\rbrack /\eta)},
\label{eq:softmax_sched}
\end{equation}

with the temperature parameter $\eta$, schedule $\cT_h$ at step $h$, auxiliary task reward $R_{\cI}$.
In comparison to the original approach the improved scheduler has a considerable influence when we deal with a larger number of auxiliaries and reward definitions that are potentially contradicting each other - or in contrast are trivial.

\subsection*{Agent Details}
For the pixel baseline experiments, we use a 3-layer convolutional policy and Q-function torsos with [64, 64, 64] feature channels, [(4, 4), (3, 3), (3, 3)] as kernels and stride 2 which we denote as standard pixel torso below.

\begin{table}[h]
\begin{center}
 \begin{tabular}{c||c} 
 Architectures \\ \hline
 Policy torso (pixels) \\(shared across tasks) & standard pixel torso\\ 
 Policy torso (features)\\(shared across tasks) & 512\\ 
 Policy heads \\ (task-dependent) & 200 \\ 
 Q function torso (pixels) \\(shared across tasks) & standard pixel torso\\
 Q function torso (features) \\(shared across tasks) & 400\\
 Q function heads \\ (task-dependent) & 300\\
 Activation function & elu\\
 Layer norm on first layer & Yes\\
 Tanh on output of layer norm & Yes\\
 Tanh on input actions to Q-function & Yes \\ \hline
 MPO parameters \\ \hline
 $\epsilon_{\mu}$ & 1e-3 \\
 $\epsilon_{\Sigma}$ & 1e-5\\
 Number of action samples& {20}\\ \hline
 General parameters \\ \hline
 Replay buffer size (pixels) & 1e6 \\
 Replay buffer size (features) & 3e6  \\
 Target network \\ update period & 500\\
 Batch size (pixels) & 64\\
 Batch size (features)& 512\\
 Discount factor ($\gamma$) & 0.99 \\
 Adam learning rate & 3e-4 \\
 Retrace sequence length & 8
\end{tabular}
\end{center}
\caption{Agent Hyperparameters }
\label{tab:agent}
\end{table}

\section*{Additional Experiments}

\begin{figure*}
\begin{minipage}[b]{\textwidth}
	\centering
	\begin{tabular}{cccc}
         & Lift & Stack & Push \\
        \hspace{-2mm}
        \rotatebox{90}{~~~~~~~Full Model} & \hspace{-3mm}
        \includegraphics[trim=0.3cm 0 0 0, clip,height = 0.17\textwidth]{figures/features/icra/RLRL_full_liftnoleg_icra.pdf} &
        \hspace{-5mm}
        \includegraphics[trim=0.3cm 0 0 0, clip,height = 0.17\textwidth]{figures/features/icra/RLRL_full_stacknoleg_icra.pdf} &
        \hspace{-5mm}
        \includegraphics[trim=0.3cm 0 0cm 0, clip,height = 0.17\textwidth]{figures/features2/RLRL_fullnoleg.pdf}\\
        \hspace{-2mm}
        \rotatebox{90}{~~~~~~Medium Model} & \hspace{-3mm}
        \includegraphics[trim=0.3cm 0 0 0, clip,height = 0.17\textwidth]{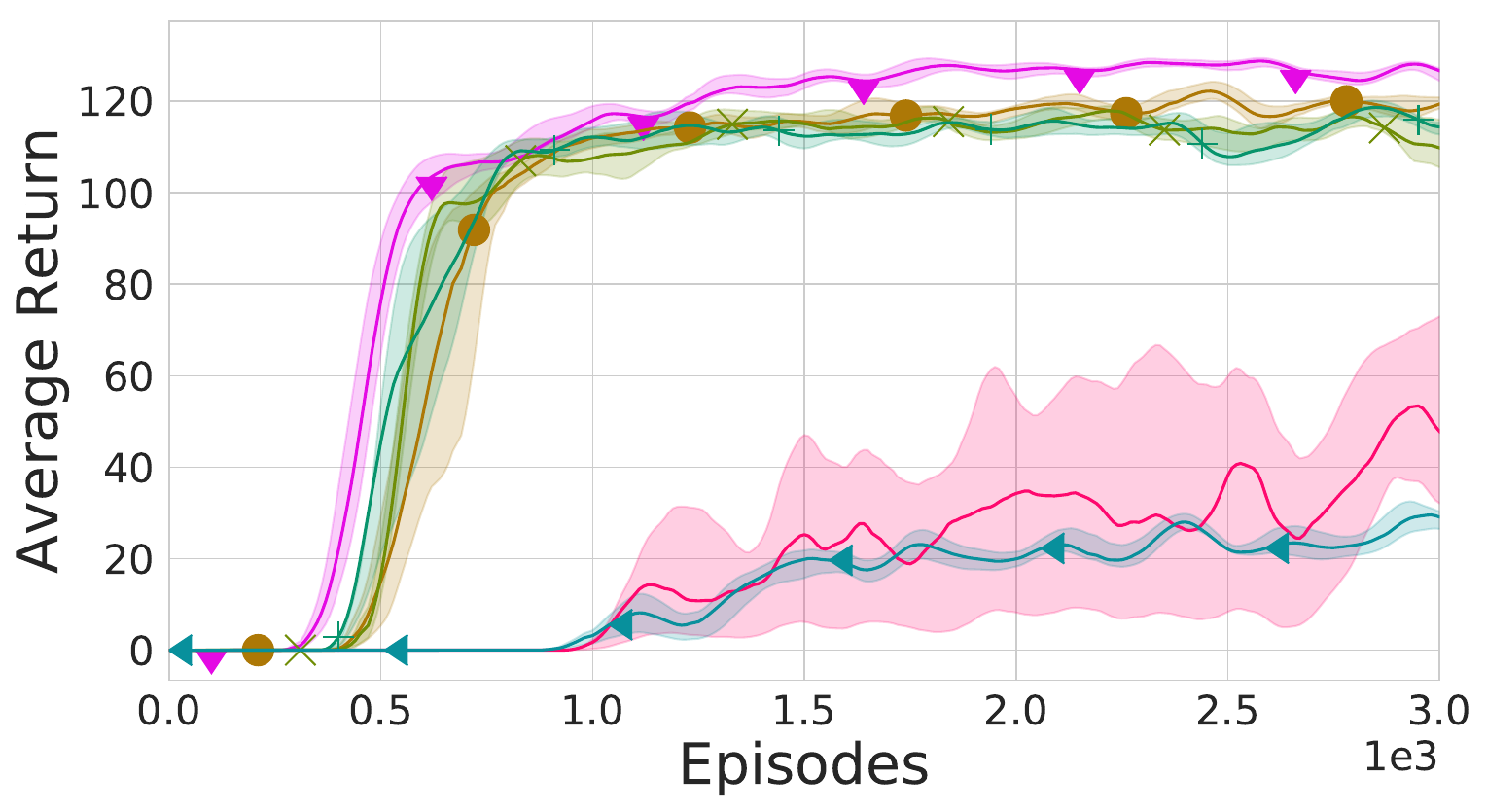} &
        \hspace{-5mm}
        \includegraphics[trim=0.3cm 0 0 0, clip,height = 0.17\textwidth]{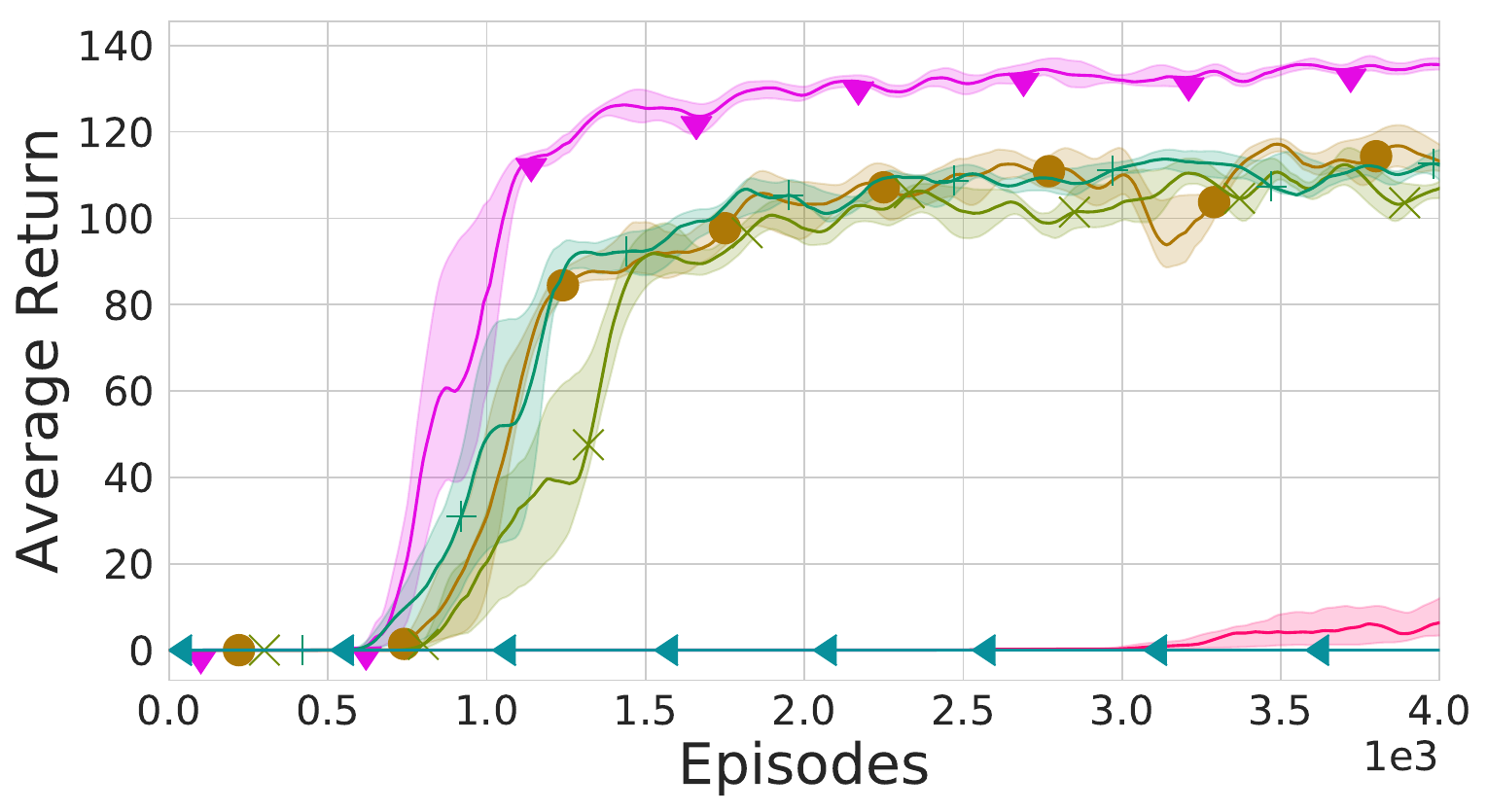}&
        \hspace{-5mm}
        \includegraphics[trim=0.3cm 0 0 0, clip,height = 0.17\textwidth]{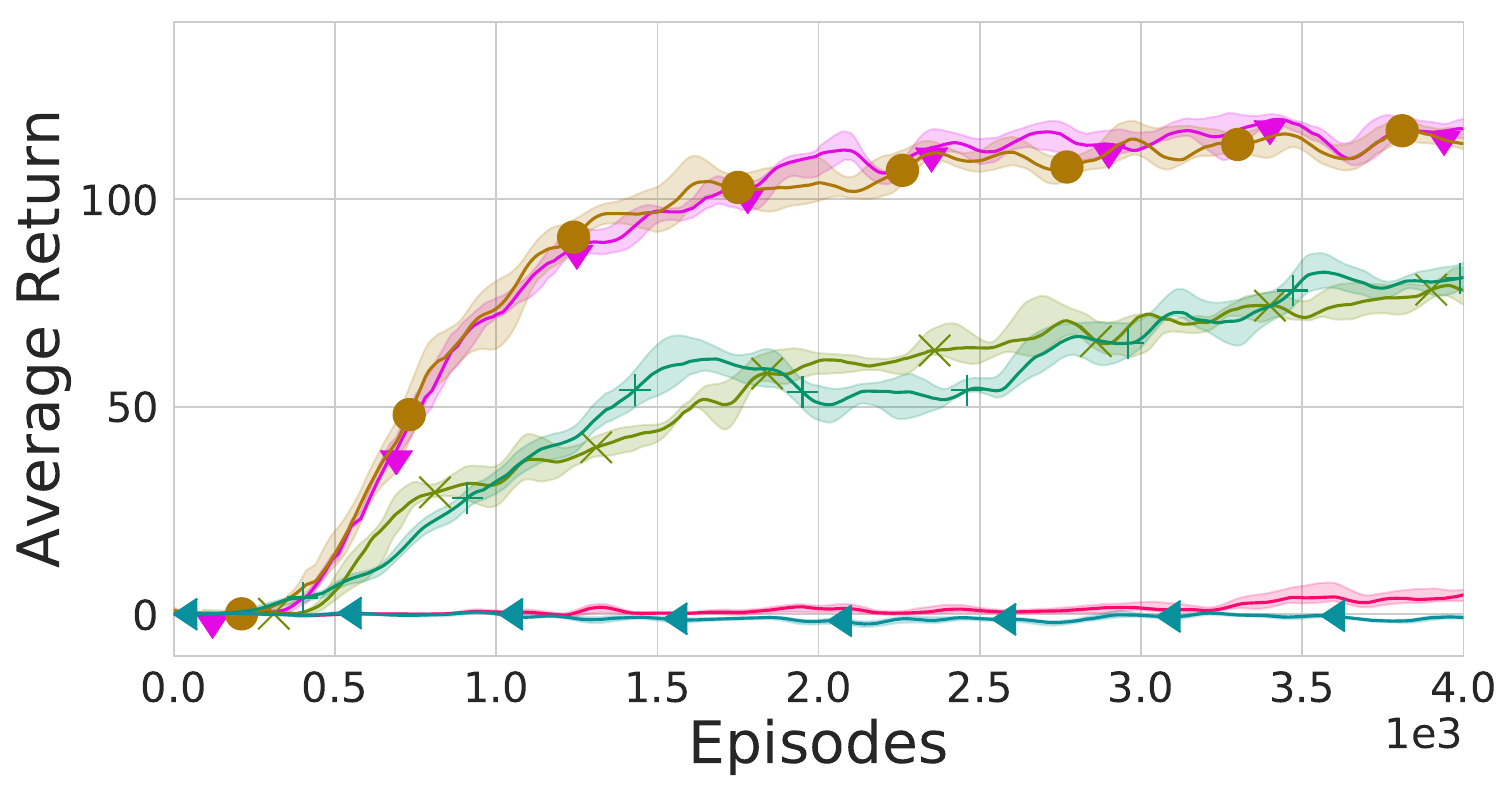}\\
        \hspace{-2mm}
        \rotatebox{90}{~~~~~~Small Model} & \hspace{-3mm}
        \includegraphics[trim=0.3cm 0 0 0, clip,height = 0.17\textwidth]{figures/features/icra/RLRL_only_green_liftnoleg_icra.pdf} &
        \hspace{-5mm}
        \includegraphics[trim=0.3cm 0 0 0, clip,height = 0.17\textwidth]{figures/features/icra/RLRL_only_green_stacknoleg_icra.pdf}&
        \hspace{-5mm}
        \includegraphics[trim=0.3cm 0 0 0, clip,height = 0.17\textwidth]{figures/features2/RLRL_only_greennoleg.pdf}
	\end{tabular}
    \caption{Complete input representation arguments including the partial representation model. Lifting (left), stacking (middle) and pushing (right) experiments for small (bottom), partial (middle) and full (top) input representations.}
	\label{fig:app_all_inputs}
\end{minipage}
\end{figure*}

\begin{figure}[h]
	\centering
	\begin{tabular}{c}
        \includegraphics[trim=0cm 0cm 0 0, clip,width = 0.45\textwidth]{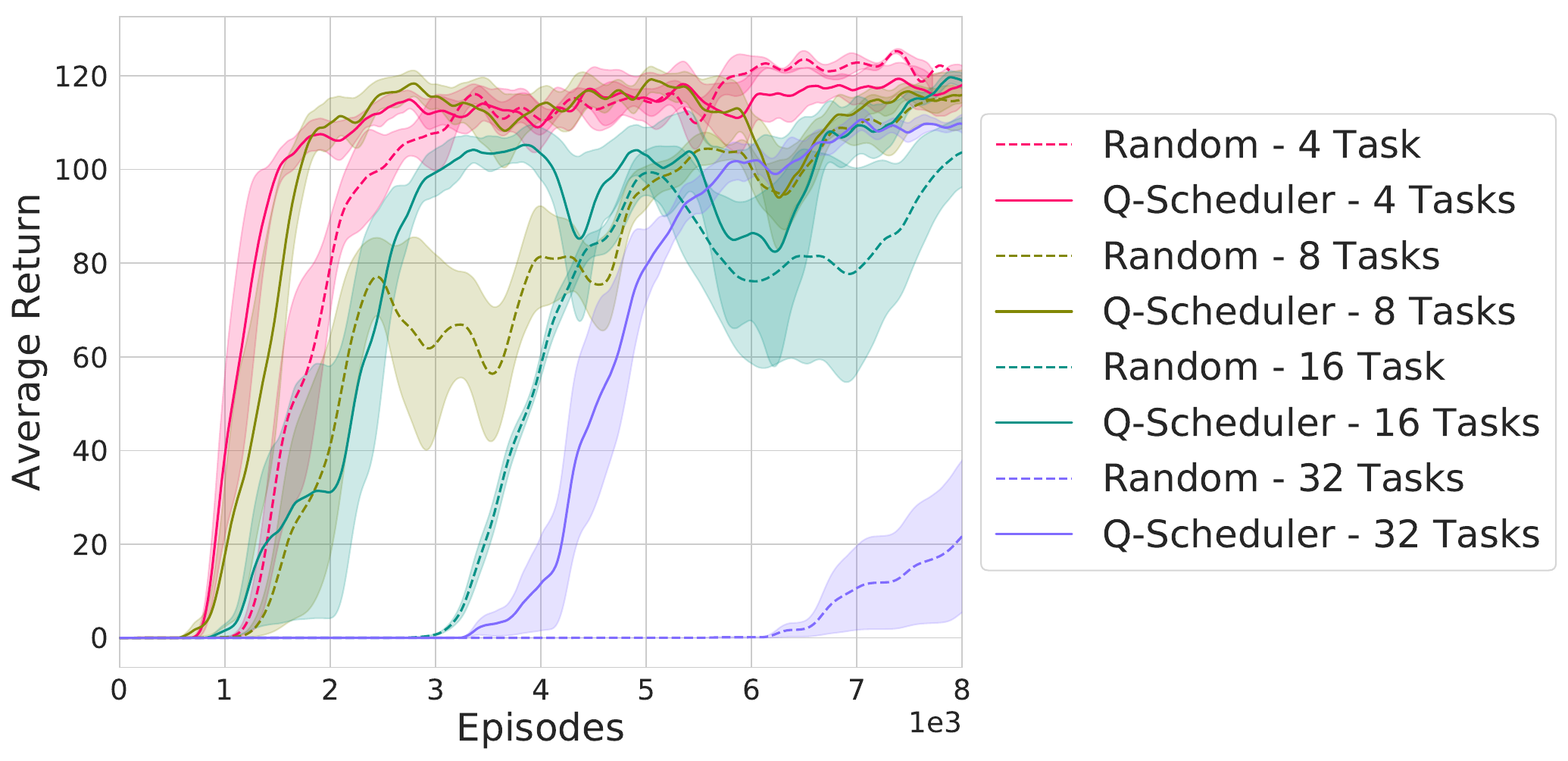}
	\end{tabular}
    \caption{Learning curves for scheduling ablations for growing number of tasks. We use  4, 8, 16, or 32 tasks from the representation modules and the main task. Increasing number of tasks strongly reduces performance and increases data requirements with random task sampling. }
	\label{fig:app_scheduling}
\end{figure}

\end{document}